\documentclass{article}

\usepackage[preprint]{neurips_2024}
\usepackage{amsmath}
\usepackage{wrapfig}
\usepackage{graphicx} 
\usepackage{lipsum}

\usepackage{graphicx} 
\usepackage[utf8]{inputenc} 
\usepackage[T1]{fontenc}    
\usepackage{hyperref}       
\usepackage{url}            
\usepackage{booktabs}       
\usepackage{amsfonts}       
\usepackage{nicefrac}       
\usepackage{microtype}      
\usepackage{xcolor}         
\usepackage{graphicx}
\usepackage{subcaption}
\usepackage{wrapfig}
\usepackage[capitalize,noabbrev]{cleveref}

\usepackage{color}
\definecolor{myfavblue}{rgb}{0.05, 0.2, 0.8}
\definecolor{keywords}{RGB}{255,0,90}
\definecolor{comments}{RGB}{0,0,113}
\definecolor{red}{RGB}{160,0,0}
\definecolor{green}{RGB}{0,150,0}
\definecolor{C0}{rgb}{0.12156862745098039, 0.4666666666666667, 0.7058823529411765}  

\definecolor{myblue}{HTML}{3182bd}
\definecolor{myred}{HTML}{de2d26}

\definecolor{mydarkblue}{rgb}{0,0.08,0.45}
\hypersetup{
    colorlinks=true,
    linkcolor=mydarkblue,
    citecolor=mydarkblue,
    filecolor=mydarkblue,
    urlcolor=mydarkblue
}

\title{Efficient Continual Pre-training  by \\ Mitigating the Stability Gap}

\author{%
   Yiduo Guo$^{1}$,~~Jie Fu$^{2}$,~~Huishuai Zhang$^{1}$,~~Dongyan Zhao$^{1}$,~~Yikang shen$^{3}$\\
    $^{1}$Peking University, $^{2}$HKUST, $^{3}$MIT-IBM Watson AI Lab
    \\
    \texttt{yiduo@stu.pku.edu.cn, 
 jiefu@ust.hk, zhanghuishuai@pku.edu.cn,} \\ 
    \texttt{zhaodongyan@pku.edu.cn, yikang.shn@gmail.com}\\
}

\begin{document}

\maketitle

\begin{abstract}
Continual pre-training has increasingly become the predominant approach for adapting Large Language Models (LLMs) to new domains. 
This process involves updating the pre-trained LLM with a corpus from a new domain, resulting in a shift in the training distribution. To study the behavior of LLMs during this shift, we measured the model's performance throughout the continual pre-training process. we observed a temporary performance drop at the beginning, followed by a recovery phase, a phenomenon known as the "stability gap," previously noted in vision models classifying new classes.  
The substantial performance drop and slow recovery associated with this gap lead to inefficient pre-training for domain performance improvement and the forgetting of general task knowledge.
To address this issue and enhance LLM performance within a fixed compute budget, we propose three effective strategies: (1) Continually pre-training the LLM on a subset with a proper size for multiple epochs, resulting in faster performance recovery than pre-training the LLM on a large corpus in a single epoch; (2) Pre-training the LLM only on high-quality sub-corpus, which rapidly boosts domain performance; and (3) Using a data mixture similar to the pre-training data to reduce distribution gap. 
{\color{black}We conduct various experiments on Llama-family models to validate the effectiveness of our strategies in both medical continual pre-training and instruction tuning. }
For example, our strategies improve the average medical task performance of the OpenLlama-3B model from 36.2\% to 40.7\% with only 40\% of the original training budget and enhance the average general task performance without causing forgetting. 
{\color{black}Furthermore, we apply our strategies to continually pre-train and instruction-tune the Llama-3-8B model. The resulting model, \textbf{Llama-3-Physician}, achieves the best medical performance among current open-source models, and performs comparably to or even better than GPT-4 on several medical benchmarks.} {We release our models at \url{https://huggingface.co/YiDuo1999/Llama-3-Physician-8B-Instruct}}.
\end{abstract}
\section{Introduction}
Continual pre-training is an important approach for LLMs to  improve their performance in target domains~\citep{huang2023lawyer,yang2024pLlama,chen2023meditron70b}, learn new topics and languages~\citep{jiang2024instruction, Gupta2023ContinualPO}, and even boost their general capabilities~\citep{ibrahim2024simple}. However, while many studies explore the mechanisms and properties of LLMs during pre-training from scratch~\citep{Biderman2023PythiaAS,xue2024repeat},  only a few research works focus on the behavior of LLMs during continual pre-training. 
To investigate the learning dynamic of LLM continual pre-training, 
we conduct experiments in the medical domain and closely monitor changes in model performance throughout the training process.
Surprisingly, we find that the LLM's performance on medical tasks drops at the early stage of training, despite a consistent improvement in perplexity on the medical corpus. However,  as training progresses and more data are used, the task performance recovers and outperforms the original model.
We further find that the initial performance drop and the following performance recovery phenomenon generally happen in LLMs across different scales and 
different training corpus.

\begin{wrapfigure}{l}{0.5\textwidth} 
  \centering
  \includegraphics[width=0.49\textwidth]{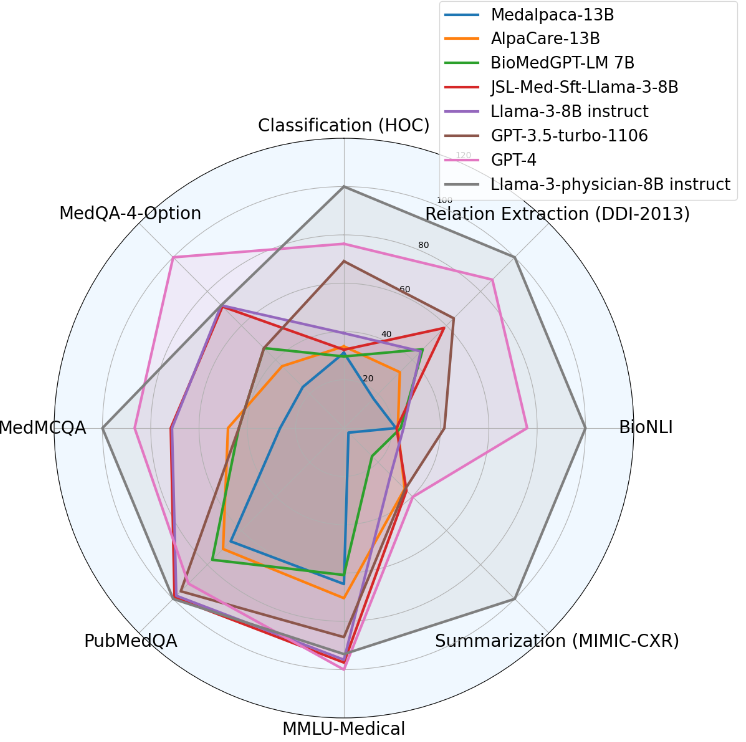} 
  \caption{The performance comparison between our model (Llama-3-physician) and other baselines involves reporting the ratio of each model's task performance to the best performance of that task among all models.}
\label{fig.compare}
\end{wrapfigure}
To explain the LLM's abnormal behavior during continual pre-training, we revisit the concept of the stability gap~\citep{de2022continual,caccia2021new} from continual learning. 
The stability gap describes the phenomenon where the performance of old tasks initially drops and then recovers when learning a new task.
A previous study~\citep{de2022continual} explains this initial drop as the model having only a small stability gradient to maintain the performance of previous tasks compared to a large plasticity gradient for learning the new task at the beginning of continual learning. 
After the initial stage, the stability gradient will rise, leading to performance recovery. 
Following this framework, we hypothesize that the LLM's performance on medical tasks depends on both medical domain knowledge and general instruction-following ability. 
Medical domain knowledge can be improved with the plasticity gradient, while the instruction-following ability is more related to the stability gradient. 
Therefore, the initial drop in medical task performance occurs due to an insufficient stability gradient to maintain the ability to follow instructions. To verify this hypothesis, we further study the model's general domain task performance during continual pre-training on our medical-related corpus.
The results show a similar v-shape curve as the medical domain tasks, where the task performance initially decreases and then starts to recover. Furthermore, from the perspective of weight updates, we find that the top layers' weights, which contain high-level task knowledge, have comparatively small weight updates only at the initial stage. This also suggests that the LLM has only a small initial stability gradient to protect its instruction-following ability. 

The stability gap causes inefficiency in continual pre-training as it delays the improvement in the LLM's performance. To address this, we propose three efficient continual pre-training strategies:
\begin{itemize}
    \item[1.]Instead of continually pre-training the LLM on a large corpus for one epoch, which induces a large plasticity gradient for a long period, we continually pre-train the LLM on a subset of the corpus with a proper size for multiple epochs.
    \item[2.] Select the subset with the highest-quality tokens to learn rich domain knowledge, leading to faster performance recovery and higher peak performance.
    \item[3.] Use a data mixture rate similar to the pre-training data, thus reducing the distribution shift and mitigating the knowledge forgetting of general instruction-following ability.
\end{itemize}

To verify our strategy, we first conduct experiments on the OpenLlama-3B model. 
We find that our strategies not only accelerate performance improvement by mitigating the stability gap but also improve the LLM's peak performance. 
We also compare our strategies with other continual pre-training techniques and analyze the influence of important learning factors, such as learning rate, for our strategies in Sec.~\ref{sec.eval}. 
{\color{black}Finally, we apply our strategies to both the continual pretraining and instruction tuning processes of the Llama-3-8B model~\citep{LLaMa-3}, efficiently enhancing its performance on diverse medical tasks, outperforming other open-source LLM baselines, and achieving performance comparable to GPT-4 (See performance comparison in Figure~\ref{fig.compare}).}

\section{Related work}
\paragraph{Large language Models} such as GPT-4~\citep{openai2023gpt4}, Gemini~\citep{teamgemini}, and Llama~\citep{touvron2023Llama}), have billions of parameters and show strong performance on various basic natural language tasks~\citep{qin2023chatgpt}, human examination~\citep{Hendrycks2020MeasuringMM,zhong2023agieval}, and agent-related tasks~\citep{guo2023pptc,liu2023agentbench,zhou2023webarena}. Their success attracts researchers to analyze LLMs' learning properties during the pre-training process~\citep{Kaplan2020ScalingLF, Biderman2023PythiaAS, Zhang2024TinyLlamaAO}. ~\citet{Kaplan2020ScalingLF} finds the pre-training scaling rule for model size and dataset size and then~\citet{Hoffmann2022TrainingCL} proposes the Chinchilla rule that claims the equal importance of the model size and the number of training tokens. ~\citet{Sorscher2022BeyondNS} further claims that pruning low-quality data can improve the above neural scaling laws. However, high-quality training tokens are limited and may be run out soon~\citep{Villalobos2022WillWR}. Thus, some researchers try to maximize the utilization of the existing corpus by training it for multiple epochs~\citep{muennighoff2024scaling,xue2024repeat}. But they observe the performance degradation~\citep{hernandez2022scaling, Xue2023ToRO, Hoffmann2022TrainingCL} after training 4 epochs.

\paragraph{Continual pre-training} gradually becomes necessary for LLMs to expand their basic ability~\citep{wu2022continued,fu2024data,zhuang2024structlm}, avoid outdated information~\citep{jiang2024instruction}, and become the domain expert~\citep{huang2023lawyer,yang2024pLlama,chen2023meditron70b,Nguyen2023AstroLlamaTS,Wu2023PMCLlamaTB,yildiz2024investigating,Xie2024MeLF}.
The domain corpus for continual pre-training can be collected by n-gram models~\citep{muennighoff2024scaling}, heuristic rules designed by human experts~\citep{chen2023meditron70b, Zhang2024AutoMathTextAD} or automatically identified by a LLM~\citep{Zhang2024AutoMathTextAD}.
For the continual pre-training techniques. ~\citet{ke2023adapting,ke2022continual} focused on adding masks or adjusting the architecture of small Language models like RoBERT to protect the learned general knowledge. However, these techniques result in huge computational consumption for LLMs.
Recent studies~\citep{Gupta2023ContinualPO} show that learning rate re-warming can improve LLMs' downstream task performance. ~\cite{ibrahim2024simple} further claims that learning rate re-warming, re-decaying, and replay can make the continual pre-training performance match the performance of fully re-training when continually pre-training the English LLM on the German corpus. Other continual pre-training method studies focus on selecting useful tokens~\citep{lin2024rho}, expanding MOE architecture~\citep{chen2023lifelong}, and knowledge distillation~\citep{jin2021lifelong}. 

\paragraph{Continual learning and the Stability Gap} Continual learning aims to design methods that can learn new knowledge without the catastrophic forgetting of previously learned knowledge~\citep{kirkpatrick2017overcoming,van2022three}. To mitigate the forgetting problem when learning a new task, replaying previous tasks' data~\citep{rolnick2019experience,buzzega2020dark,prabhu2020gdumb,buzzega2021rethinking,guo2022online} becomes the main approach. ~\citet{de2022continual,caccia2021new} further find that, although they conduct the replay approach, the vision model still first loses its performance stability in previous classification tasks ( the performance drops abruptly) and then gradually recovers. They call it the stability gap phenomenon. Different from them, we focus on the continual pre-training of the LLM and observe that both the LLM's domain task performance and general ability suffer from the stability gap. 

\section{Identifying the stability gap in continual pre-training}
In this section, we first study the behavior of large language models (LLMs) during continual pre-training by measuring them at regular intervals. We observe that the performance on the target task initially drops and then rises during continual pre-training. To explain our observations, we introduce the concept of the stability gap to continual pre-training and verify our explanations with experiments.
\subsection{Investigating the behavior of LLMs during continual pre-training}
\label{sec.identify}
\paragraph{Experiment setup}
In this study, we chose OpenLlama3B-v2~\citep{openlm2023openLlama} as our default LLM and use the medical domain as our primary target domain. Following previous work~\citep{chen2023meditron}, we set the compute budget to 50 billion (50B) training tokens. To collect the continual pre-training corpus, we follow the simple and scalable methodology of~\cite{muennighoff2024scaling}. First, we train a KenLM model~\citep{Heafield2011KenLMFA} on a high-quality medical reference corpus. Then, we use the trained KenLM model to calculate the perplexity (PPL) of samples in the Refined-Web dataset~\citep{penedo2023refinedweb}. Finally, we extract 50B tokens from the Refined-Web dataset with the lowest PPL to create the medical corpus. More details are provided in Appendix~\ref{appendix.train_details_1}.
\label{sec:identify}
\begin{figure*}[htp]
\centering 
\includegraphics[height=0.32\textwidth,width=1\textwidth]{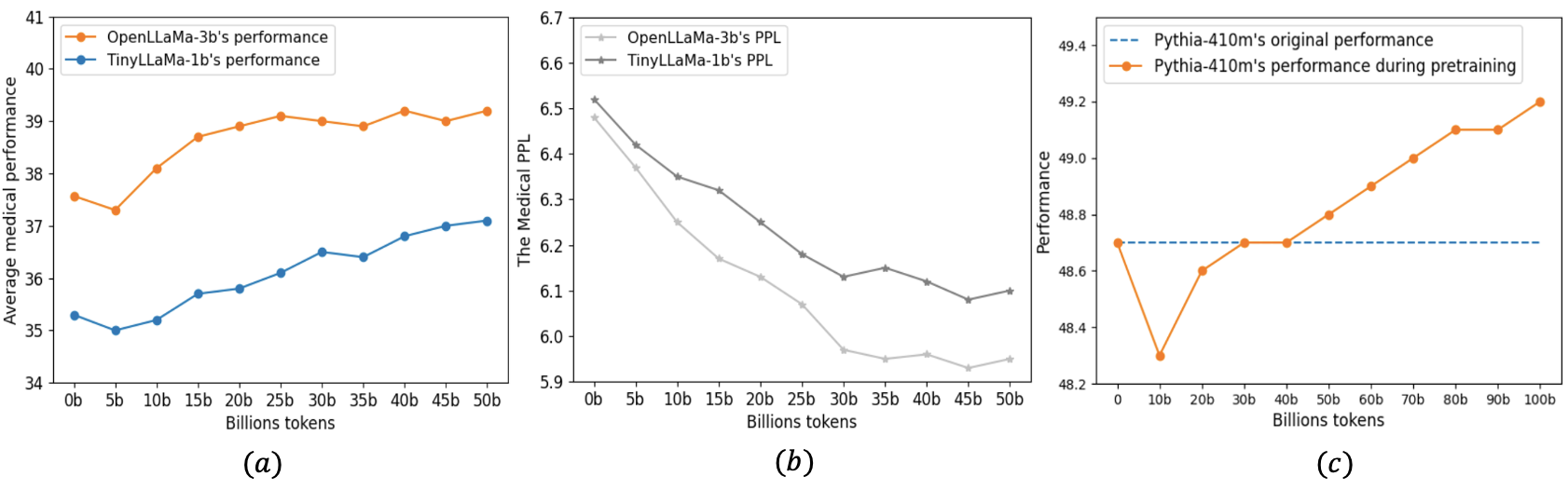} 
\vspace{-3mm}
\caption{(a) reports the models' average medical performance during the medical continual pre-training process. (b) illustrates the models' average medical perplexity (PPL) during the medical continual pre-training process. (c) shows the Pythia model's average common-sense task performance when we continually pre-train it on the new Refined-Web datasets. 
}
\label{Fig.analysis_1}
\vspace{-3mm}
\end{figure*}

\paragraph{Observation (1): The medical task performance first drops and rises during continual pre-training.} 
Specifically, we follow~\cite{chen2023meditron70b} and measure the average accuracy performance over the MMLU-Medical-Genetics~\citep{hendrycks2020measuring}, MedQA~\citep{jin2021disease}, PubMedQA~\citep{jin2019pubmedqa}, and MedMCQA~\citep{pmlr-v174-pal22a} tasks (see task details in Appendix~\ref{appendix.task}). We report the average performance on medical tasks every 5 billion training tokens. From Figure~\ref{Fig.analysis_1}(a), we observe that the domain task performance initially drops during the first 5 billion tokens and then gradually recovers and improves. Additionally, we consider the TinyLlama model~\citep{zhang2024tinyLlama}, a 1.1B Llama model trained on 3 trillion tokens, and continually pre-train it on the medical corpus. From Figure~\ref{Fig.analysis_1}(a), we observe that its performance on medical tasks also shows the same trend, despite being trained on so many tokens. 

\paragraph{Observation (2): The perplexity of medical Wikipedia steadily declines during continual pre-training.} 
We further measure the average perplexity (PPL) of the models on the Wikipedia corpus about medical terms\footnote{\url{https://huggingface.co/datasets/gamino/wiki_medical_terms}}. From Figure~\ref{Fig.analysis_1}(b), we observe that the PPL steadily drops. This indicates that the LLM has acquired medical domain knowledge at the initial continual pre-training and continues improving its medical domain knowledge throughout the entire continual pre-training process.

\paragraph{Observation (3): The general task performance also first drops and then rises during general continual pre-training.} 
Continual pretraining on another large corpus is an important approach to boost the pretrained LLM's general task performance~\citep{jiang2024instruction, Gupta2023ContinualPO}. We call it the general continual pretraining setting. We further find that it also exists a similar performance phenomenon. Specifically, we continually pre-train the Pythia-410m model~\citep{biderman2023pythia} (initially pre-trained on the Pile~\citep{Gao2020ThePA} dataset) on the RefinedWeb dataset~\citep{penedo2023refinedweb} to boost its general ability. We measure its general ability using the average performance across 10 common-sense tasks and report the average performance of every 10 billion tokens. Training details are in Appendix~\ref{appendix.train_details_1} and task details are in Appendix~\ref{appendix.task}. From Figure~\ref{Fig.analysis_1}(c), we observe that the LLM's general task performance first drops significantly and then gradually rises.

Based on our observations, the initial drop followed by a rise in target task performance is a general phenomenon in the continual pre-training of LLMs of various sizes. This abnormal behavior can be explained with the concept of the stability gap~\citep{DeLange2022ContinualEF} in the following section.
\begin{figure*}[htp]
\centering 
\includegraphics[height=0.35\textwidth,width=0.85\textwidth]{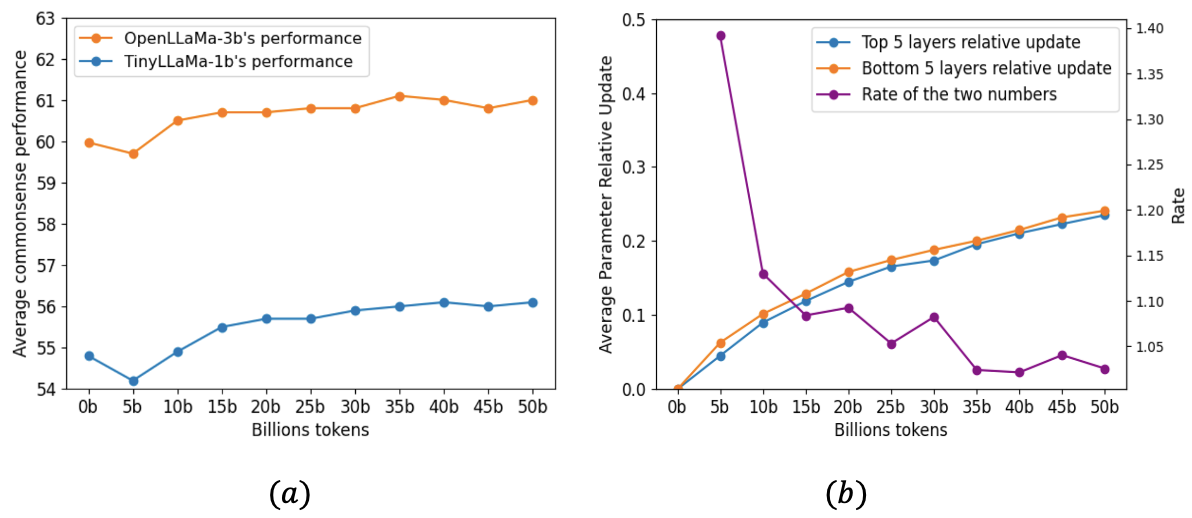} 
\vspace{-3mm}
\caption{(a) shows the OpenLLaMa's average common-sense task performance during medical continual pre-training. (b) illustrates the OpenLlama model's relative parameter update during the medical continual pre-training process. We report the average weight relative update of weights in the top 5 layers and the bottom 5 layers. We also report the rate between the two average numbers. 
}
\label{Fig.analysis_6}
\vspace{-3mm}
\end{figure*}
\subsection{Stability Gap: A conceptual explanation for the initial performance drop and then following recovery.}
\paragraph{The stability gap} phenomenon has been observed as the model's performance in previous tasks first drops and then rises when learning the new task and ~\citet{DeLange2022ContinualEF} explains it by disentangling the gradient $\mathcal{L}$ in $\alpha$-weighed plasticity and stability gradients: $\mathcal{L}=\alpha\mathcal{L}_{\textit{plasticity}}+(1-\alpha)\mathcal{L}_{\textit{stability}}$ where $\mathcal{L}_{\textit{plasticity}}$ aims to learn to the new task and $\mathcal{L}_{\textit{stability}}$ maintains the previous tasks' performance. 
The performance of previous tasks initially drops because the plasticity gradient exceeds the stability gradient, causing neglect in maintaining prior task performance. Subsequently, the performance loss in previous tasks enhances the stability gradient, while adaptation to the new task reduces the plasticity gradient, leading to gradient balance and performance recovery.

\paragraph{Explanation of our observations} We assume that the performance on the new domain task relies on both the target domain knowledge and the instruction-following ability during continual pre-training. Inspired by the stability gap explanation framework, we decouple the continual pre-training gradient into the plasticity gradient for learning new domain knowledge and the stability gradient for maintaining instruction-following ability. Initially, the stability gradient is small compared to the plasticity gradient, causing a performance drop on the target domain because of the destructed instruction-following ability.   Subsequently, the stability gradient increases to recover the instruction-following ability, while the plasticity gradient already helps the LLM learn domain knowledge, observing performance improvement on the target domain.  

\paragraph{Empirical verification for our explanation} We verify our explanation by (1) measuring general task performance during continual pre-training on the medical-related corpus. As shown in Figure~\ref{Fig.analysis_6}(a), the general task performance follows a similar V-shape curve, indicating the recovery of general instruction-following ability after the initial drop.  We also verify our explanation at the weight level by (2) measuring the relative weight update of each weight $w$ as $\frac{w_{\textit{t}}-w_0}{w_0}$, where $w_{\textit{t}}$ is the weight value during continual pre-training and $w_0$ is the original weight value. A high relative weight update indicates a large gradient for updating the weight. Figure~\ref{Fig.analysis_6}(b) shows that the bottom layers' weights initially have a higher relative weight update than the top layers (rate > 1.35). Previous studies indicate that bottom layers learn the syntax and low-level semantics~\citep{Devlin2019BERTPO, Hewitt2019ASP, Ling2023DomainSA}, while top layers contain high-level semantics and task-specific knowledge~\citep{Yang2024LaCoLL, Chen2024CompressingLL}. This suggests that the top layers' weights indeed lack sufficient stability gradient to maintain instruction-following ability initially. The performance then recovers as the relative weight updates (stability gradient) increase in the top layers and domain knowledge is learned, as indicated by the continuous drop in medical perplexity. 

\section{Efficient continual pretraining strategies for mitigating the stability gap}
\label{sec.method}
In this section, we propose three efficient continual pre-training strategies for overcoming the above stability gap problem. The training process and details follow those in the above section.

\begin{figure*}[htp]
\centering 
\includegraphics[height=0.3\textwidth,width=1.015\textwidth]{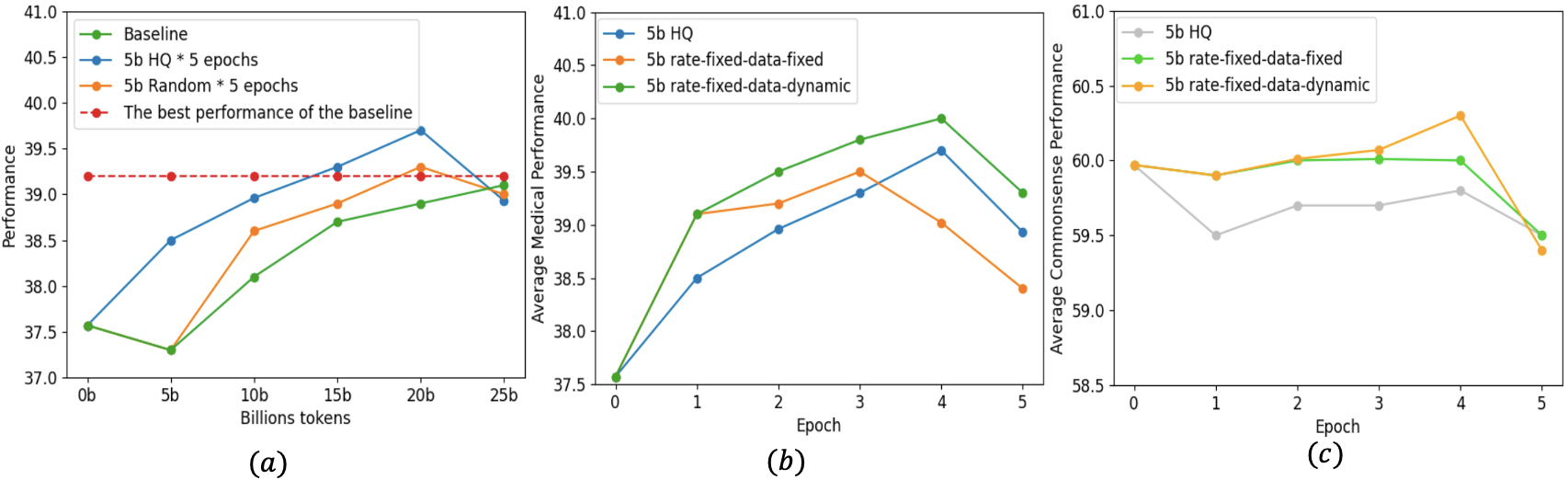} 
\vspace{-7mm}
\caption{(a) reports the average medical performance during the medical continual pre-training process. The baseline is pre-training the OpenLlama-3B model with 50b medical tokens with one epoch. '5b Random' is pre-training the LLM with 5b tokens randomly selected from the 50b medical tokens for 5 epochs. '5b HQ' is pre-training the LLM with the highest quality (HQ) 5b tokens of the 50b medical tokens for 5 epochs. (b) shows the average medical performance across 5 epochs. (c) illustrates the average commonsense task performance across 5 epochs. 
}
\label{Fig.analysis_2}
\vspace{-3mm}
\end{figure*}
\paragraph{Strategy I: Continually pre-train the LLM on a corpus subset across multiple epochs rather than the entire large corpus for a single epoch.}
The domain task performance depends on the general instruction-following ability, which is related to the rise of the stability gradient. The usual continual pre-training approach collects as many samples as possible and trains them one time. However, this approach means the LLM maintains a high plasticity gradient for learning new samples in every batch, causing a stability gap and a slow rise in the stability gradient. To mitigate this, we propose randomly selecting a subset with a proper size from the entire corpus and pre-training the LLM on this subset across multiple epochs. This method reduces the need for a high plasticity gradient after the first epoch and accelerates the rise of the stability gradient (performance recovery). In Figure \ref{Fig.analysis_2}(a), we observe that this strategy leads to faster performance recovery. The LLM achieves peak performance at the fourth epoch, consistent with previous studies~\citep{xue2024repeat}.

\paragraph{Strategy II: Continually pre-train the LLM on the corpus subset with the highest quality.} 
The performance of domain tasks also depends on the learned domain knowledge. Therefore, collecting a subset with the highest quality should further enhance performance. To verify this, we used the trained KenLM from Sec.~\ref{sec.identify} to calculate the perplexity (ppl) of each sample in the entire medical corpus. A lower perplexity indicates that the sample is closer to the distribution of the medical reference corpus. We then continually pre-trained the OpenLlama-3B model on the subset with the lowest perplexity (i.e., the highest quality) for multiple epochs. From Figure~\ref{Fig.analysis_2} (a), we observe that the high-quality subset indeed enables the LLM to recover performance faster and stronger in the medical domain. Further analysis of the pre-training subset size is presented in Sec.~\ref{sec.analysis}.

\paragraph{Strategy III: Use a data mixture rate similar to the pre-training data.} 
The pre-training data mixture rate is a vital factor for the pre-training performance of large language models (LLMs)~\citep{xie2024doremi,shen2023slimpajama}. Therefore, we propose a third strategy that follows the pre-training data's mixture rate to construct the continual pretraining training subset, aiming to reduce the distribution gap and stabilize the instruction-following ability of the LLM during continual pre-training. Specifically, for the OpenLlama model, we follow the Llama mixture rate~\citep{touvron2023Llama} to collect 5 billion tokens initially. We then replace the CC and C4 data (82\% of the 5 billion tokens) with medical tokens sampled from the highest quality 5 billion medical tokens (HQ-5b). There are two ways to sample these medical tokens. The first method randomly samples the medical tokens once to construct a fixed training corpus. We call this ``rate-fixed-data-fixed''. The second method randomly samples the medical tokens from the HQ-5b tokens for each epoch. We call this ``rate-fixed-data-dynamic''.

From Figure~\ref{Fig.analysis_2}(b), we observe that both methods improve the performance of the first epoch by overcoming the stability gap. The second method achieves a higher peak performance as it offers a better trade-off between recovering the instruction-following ability through data replay and learning domain knowledge from a large number of medical tokens. Additionally, our strategies further improve the average performance on general commonsense tasks, as shown in Figure~\ref{Fig.analysis_2}(c), and reduce the medical perplexity and the rate of relative weight update, as detailed in Appendix~\ref{appendix:ppl}. We also investigate the effectiveness of our three strategies in the general continual pre-training setting in Appendix~\ref{appendix.general}.
\label{headings}

\section{Evaluation}
\label{sec.eval}
In this section, we first compare the effectiveness of our strategies with other continual pre-training techniques. Next, we investigate the impact of important learning factors, such as the learning rate, on our strategies. Finally, we deploy our strategies into the newest Llama-3-8b model, which achieves the strongest fine-tuned performance among open-source baselines.

\subsection{Comparison with other continual pre-training techniques}

\paragraph{Baselines and evaluation tasks} 
We consider the following baselines for comparison: (1) Continually pre-training the OpenLLaMa-3B LLM with 50 billion collected medical tokens for one epoch ("the full token baseline").
(2) Re-warming and re-decaying the learning rate of (1) based on the paper by~\citep{ibrahim2024simple}.
(3) Replay baselines: Following~\citep{chen2023meditron}, we replace 10\% of the tokens in (1) with tokens randomly sampled from the OpenLLaMa-3B's pre-training dataset (the RefinedWeb dataset). We also consider a replay baseline that replaces 10\% of medical tokens in the experiment using strategies I and II with randomly selected tokens from the RefinedWeb dataset. This baseline does not consider the data mixture rate. 
(4) Parameter protection baselines: Following~\citep{harun2023overcoming}, we freeze the top 5 layers' weights during the continual pre-training process of (1) to protect the high-level instruction-following ability and mitigate the stability gap. We also consider another baseline that freezes the bottom 5 layers' weights for comparison. We follow~\citep{chen2023meditron} and consider the tasks of PubMedQA, MedMCQA, and MedQA-4-Option. For the MMLU benchmark~\citep{hendrycks2020measuring}, we consider the average performance of its medical topics, including medical genetics, anatomy, clinical knowledge, professional medicine, and college medicine. We use the lm-evaluation-harness framework~\citep{eval-harness} to measure the baselines' zero-shot performance.
\begin{table}[h]
    \centering
    \resizebox{\textwidth}{!}{
        {\Huge 
        \begin{tabular}{lcccccc}
            \toprule
            \textbf{Method} &\textbf{Training tokens number} &\textbf{MMLU-Med-Avg} & \textbf{PubMedQA} & \textbf{MedMCQA} & \textbf{MedQA-4-Option} & \textbf{Avg} \\
            \midrule
            OpenLLaMa-3B & - &25.6&68.4&25.4&25.4&36.2\\
            Full token baseline &50B & 26.1 & 70.4 & 26.1 & 27.1 & 37.4 \\
            Re-warming and re-decaying &50B & 26.5 & 70.3 & 27.1 & 27.1 & 37.7 \\
            Replay 10\% data&50B & 26.3 & 69.2 & 27.6 & 26.9 & 37.5 \\
            Replay 10\% data with our strategies&20B & 29.3 & 71.0 & 30.4 & 27.6 & 38.5 \\
            Freezing top 5 layers&50B&26.2&69.9&27.1&27.3&37.6\\
            Freezing bottom 5 layers&50B&26.0&69.1&25.4&25.7&36.5 \\
            Our strategies&20B &\bf30.0&\bf71.2&\bf34.0&\bf27.8&\bf40.7\\
            \bottomrule
        \end{tabular}
        }
    }
    \caption{Zero-shot accuracy across various medical benchmarks.}
    \label{tab:pre-trained_acc}
\end{table}

\paragraph{Results} 
From Table~\ref{tab:pre-trained_acc}, we find that (1) our strategies improve the base model's average medical task performance significantly (4.5\%) with only 20 billion training tokens. This demonstrates the effectiveness and efficiency of our strategies for continual pre-training. (2) Other techniques can also improve continual pre-training performance, except for the baseline 'Freezing bottom 5 layers,' which hinders the learning of medical domain knowledge. (3) Following the pre-training data mixture rate to replay pre-training data is more effective than randomly sampling pre-training data for replay. This is because it further reduces the distribution shift between the pre-training corpus and the continual pre-training corpus, thereby helping to recover the LLM's general instruction-following ability.  
\subsection{Analysis}
\label{sec.analysis}
\begin{figure*}[htp]
\centering 
\includegraphics[height=0.3\textwidth,width=1.005\textwidth]{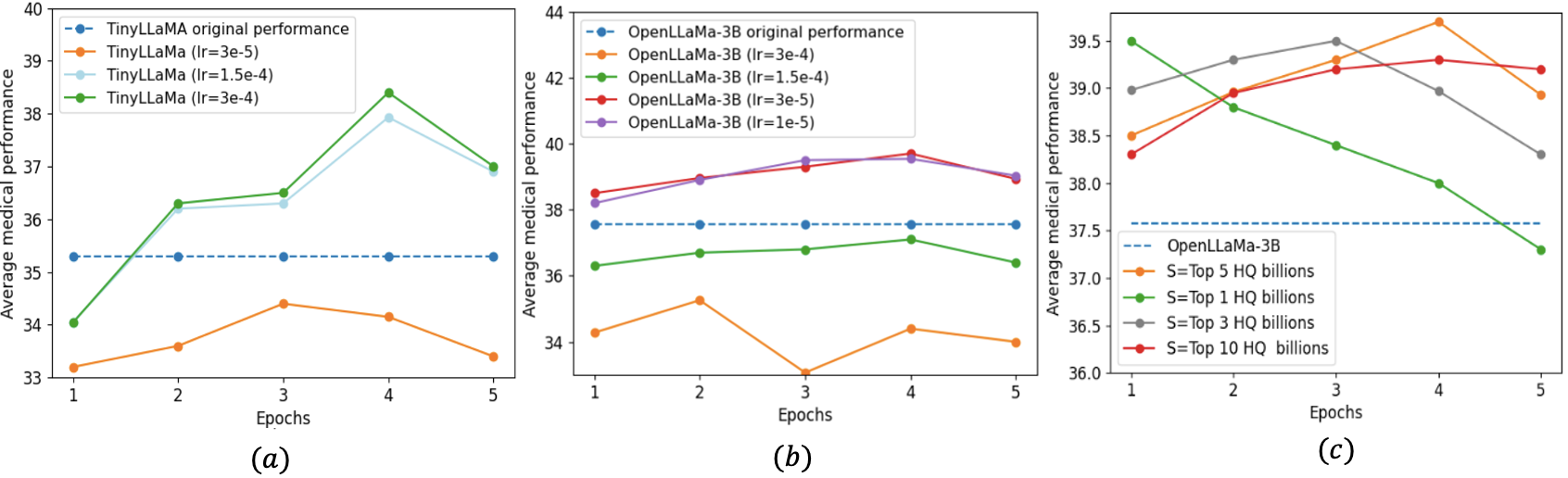} 
\vspace{-7mm}
\caption{(a) reports the performance of TinyLlama-1.1B across multiple epochs. All these experiments use our strategies with different pre-training learning rates. (b) reports the performance of OpenLlama-3B across multiple epochs. All of the experiments in (a) and (b) use our strategies with different pre-training learning rates. (c) reports the performance of OpenLlama-3B across multiple epochs with different training subset sizes $S$. To collect the pre-training corpus with different sizes, we first rank all samples of the 50 billion medical tokens based on the perplexity calculated by the trained KenLM (see Sec.~\ref{sec.identify}). Then, we select the first $S$ billion tokens with the lowest perplexity. For all experiments here, we report the average task performance of PubMedQA, MedMCQA, MMLU-medical-genetics, and MedQA-4-Option tasks.
}
\label{Fig.analysis_3}
\vspace{-3mm}
\end{figure*}
\paragraph{Impact of learning rate}
The pre-training learning rate is a crucial factor for updating LLMs during continual pre-training. To investigate its impact on our strategies, we conduct continual pre-training experiments with different learning rates. From Figure~\ref{Fig.analysis_3}(a) and (b), we find that the optimal learning rate varies with the LLM scale: a small LLM (e.g., TinyLlama-1.1B) requires a higher learning rate (e.g., 3e-4), whereas larger LLMs (e.g., OpenLlama-3B) benefit from a lower learning rate (e.g., 3e-5). If the learning rate is too low (e.g., 3e-5 for TinyLlama-1.1B), the LLM cannot learn domain knowledge effectively to boost performance. Conversely, if the learning rate is too high (e.g., 3e-4 for OpenLlama-3B), performance declines as the large learning rate leads to a significant plasticity gradient, causing the LLM to lose its general instruction-following ability for completing tasks. Based on our analysis experiments, we set the pre-training learning rate at 3e-4 for TinyLlama and 3e-5 for OpenLlama-3B's experiments.

\paragraph{Impact of the training subset size}
The size of the training subset is another important factor in our strategies. To determine the optimal training subset size, we conduct pre-training experiments on Llama-3b using various training subset sizes. From Figure~\ref{Fig.analysis_3}(c), we observe that a smaller high-quality subset yields better initial performance and mitigates the stability gap (e.g., 1 billion tokens), but it also causes the performance to drop quickly in later epochs due to overfitting. A larger subset (e.g., 10 billion tokens) results in a stability gap and slower performance recovery, as the LLM needs to maintain a high plasticity gradient to learn a large number of new samples. Based on our experiments, we select a subset with 5 billion high-quality tokens, as it mitigates the stability gap, achieves the best peak performance, and is computationally effective.
\subsection{Deploying our strategies into the Llama-3 Model}
\paragraph{Continual pre-training} 
We continually pre-train the Llama3-8B-base model using our three strategies with the high-quality 5 billion medical tokens constructed in Sec.~\ref{sec.method} for 4 epochs. The training details are in Appendix~\ref{appendix.llama-3}. After the continual pre-training process, we find that the average medical performance drops slightly, likely due to the unknown data mixture rate of Llama-3 and the lack of access to its high-quality pre-training corpus for performance recovery. However, the medical perplexity is significantly lower than that of the Llama3-8B-base model.

\paragraph{Task-specific fine-tuning} 
To evaluate LLMs' performance in the supervised learning setting, we follow~\citep{chen2023meditron} and individually conduct task-specific finetuning on both the base models and the continually pre-trained models using each benchmark’s training set. Since MMLU~\citep{hendrycks2020measuring} does not have a training set, we follow~\citep{chen2023meditron} and primarily consider the MMLU-Medical-Genetics benchmark, evaluating the model finetuned on MedMCQA.  We put task details in Appendix~\ref{appendix.task} and training details in Appendix~\ref{appendix.llama-3}.

\paragraph{Baselines} 
For task-specific fine-tuning, we consider three kinds of baselines here: (1) Task-specific finetuning of the base model of open-source LLMs. This includes models such as Llama-2-70B, Llama-3-8B, and Llama3-Aloe-8B-Alpha~\citep{gururajan2024aloe}. We copy their results from their respective papers~\citep{gururajan2024aloe} or the Meditron paper~\citep{chen2023meditron} except for the Llama-3-8B, which we finetuned using the same process as our strategies. (2) Task-specific finetuning of continually pre-trained LLMs like meditron~\citep{chen2023meditron}, BioMistral SLERP 7B~\citep{labrak2024biomistral}, Llama-3-8B-full. These LLMs have been continually pre-trained with a medical corpus. We copy their results from their papers, except for Llama-3-8B-full, for which we continually pre-train the Llama-3-8B with 50B medical tokens collected in Section~\ref{sec.identify}, and then finetune it using the same process as our strategies. (3) Closed-source LLMs. This includes models like ChatGPT and GPT-4 \citep{openai2023gpt4}. The results are measured using the Microsoft Azure OpenAI API service \citep{Shi2024MedAdapterET}.

\begin{table}[h]
    \centering
    \resizebox{\textwidth}{!}{
        \begin{tabular}{lccccc}
            \toprule
            \textbf{Model} & \textbf{MMLU-Medical} & \textbf{PubMedQA} & \textbf{MedMCQA} & \textbf{MedQA-4-Option} & \textbf{Avg} \\
            \midrule  
            Llama-2-7B~\citep{Touvron2023Llama2O} & 56.3 & 61.8 & 54.4 & 49.6 & 53.2 \\
            
             BioMistral SLERP 7B~\citep{labrak2024biomistral} &60.5&75.2&44.2&47.3&56.8 \\
    \text{MEDI\textsc{Tron}-7B}~\citep{chen2023meditron} & 55.6 & 74.4 & 59.2 & 52.0 & \text{57.5} \\
            Llama3-Aloe-8B-Alpha~\citep{gururajan2024aloe} &72.7&77.2&59.0&62.3&67.8\\
            \midrule
            Llama-2-70B & 74.7 & 78.0 & 62.7 & 61.3 & 67.2 \\
            \text{MEDI\textsc{Tron}-70B} & 73.6 & \bf80.0 & 65.1 & \bf65.4 & 69.0 \\
            \midrule  
            GPT-3.5-turbo-finetuned~\citep{Shi2024MedAdapterET} &70.5 &71.4&61.8& 63.3&66.7\\  
            \midrule
            Llama-3-8B Fine-tuned (ours) & 82.3 & 75.8 & 60.0 & 61.1 & 69.8 \\
            Llama-3-8B Full (ours) & 82.0 & 78.6 & 61.8 & 60.8 & 70.8 \\
            Llama-3-Physician-8B (ours) & \bf85.0 & 79.1 & \bf81.4 & 61.5 & \bf76.7 \\
            \bottomrule
        \end{tabular}
    }
    \caption{Accuracy comparison across various medical benchmarks in the task-specific fine-tuning setting. 
    Llama-3-8B Fine-tuned is directly fine-tuned on these tasks. 
    For 'Llama-3-8B Full', we first continually pre-trained the Llama with 50B medical tokens and then finetuned the pretrained model on these tasks.
    For Llama-3-Physician-8B, we first continually pre-trained the Llama with with our strategies and then finetuned the pretrained model on these tasks.}
    \label{tab:finetune_acc}
    \vspace{-3mm}
\end{table}

\paragraph{Results} 
We use the lm-eval-harness~\citep{eval-harness} to evaluate our model (Llama-3-Physician) and related baselines' performance. No demonstration examples are used. From Table~\ref{tab:finetune_acc}, we find that our model outperforms other baselines with similar model scales on the four evaluation benchmarks by a clear margin. This is due to the following reasons: (1) we use the newest and strongest open-source Llama-3 model rather than older Llama-2 or Mistral-7B, (2) we continually pre-train the base model with high-quality medical tokens (compared to 'Llama-3-8B fine-tuned and Llama-3-8B instruct'), and (3) our strategies further boost the gains from continual pre-training markedly (compared to 'Llama-3-8B Full'). Our model also outperforms many larger LLMs (70B) on average, meaning that users can obtain higher-quality medical services with a faster inference rate and less memory consumption. 

\subsection{Deploying our strategies into the instruction tuning process}

Instruction-tuning is an important approach to boost the LLM's performance among multiple tasks. We follow~\cite{xie2024me} and consider the instruction-tuning setting that tunes the continual pretrained Llama-3-8B model (see the above section) with a combination of question-answering tasks like PubMedQA~\citep{jin2019pubmedqa}, classification tasks like HOC~\citep{baker2016automatic}, relation extract tasks like DDI2013~\citep{segura2013semeval}, inference tasks like BioNLI~\citep{bastan2022bionli}, and summarization tasks like MIMIC-CXR~\citep{johnson2019mimic}. The specific dataset details are in Appendix~\ref{appendix.task}. Unlike the above task-specific fine-tuning, we only tune one LLM here and use the instruction-tuned LLM to test all benchmarks. We tune the Llama model for 3 epochs with a learning rate of 3e-5. More training details are in Appendix~\ref{appendix.llama-3}.
\begin{figure*}[htp]
\centering 
\includegraphics[height=0.4\textwidth,width=0.95\textwidth]{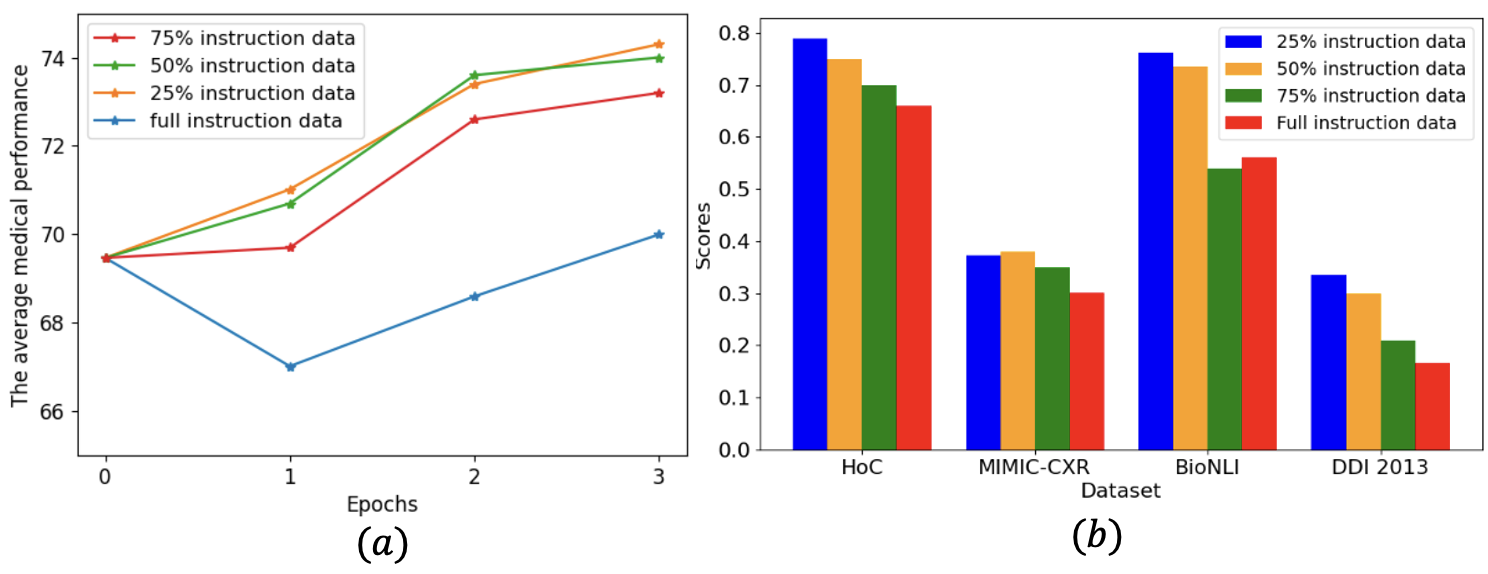} 
\vspace{-3mm}
\caption{We consider the 'full instruction data' experiment as fine-tuning the model with all instruction data for 3 epochs. For the '$n$\% data' experiments, we first uniformly sampled the highest quality instructions from each instruction dataset based on scores provided by the Deita data selector. We then mixed the sampled data with the general instructions from the Airoboros-3.2 dataset. The total training tokens are equal to $n$\% of the full instruction data. We set $n$ to 25, 50, and 75 here. (a) shows the experiments' average medical question-answering task performance during instruction tuning. (b) illustrates the experiments' performance for other medical tasks. For BioNLI, DDI 2023, and HOC tasks, we report macro-F1 as the score. For MIMIC-CXR summarization tasks, we report Rouge-L as the score.
}
\label{Fig.analysis_7}
\vspace{-3mm}
\end{figure*}
\paragraph{Deployment} In the instruction tuning process, our first strategy is common as the medical instruction tuning process usually involves multi-epochs training~\citep{zhang2023biomedgpt,xie2024me,han2023medalpaca}. For the second strategy, we consider Deita~\citep{liu2024what}, a simple automatic instruction data selector, to select high-quality medical instruction data. This selector uses the LLM to give quality scores for instructions and considers the diversity of instruction data by sampling data from different clustering. For the last strategy, we consider high-quality general instruction datasets like Airoboros-3.2~\cite {airoboros} to mitigate the forgetting in general instruction following ability. 

\paragraph{Observations} From Figure~\ref{Fig.analysis_7}, we first observe that the average performance of medical question-answering tasks initially drops slightly (in the first epoch) and then gradually rises, which is similar to the phenomenon observed in the continual pre-training process. Additionally, we observe that our strategies can mitigate the initial performance drop and achieve higher peak performance during the instruction tuning process, thereby extending the application of our strategies. Figure~\ref{Fig.analysis_7} also shows that we only need computation equivalent to 25\% of the original instruction data (consisting of high-quality medical instruction data and general instruction data) to achieve the best performance among diverse tasks. This reduces computational consumption and improves the efficiency of the instruction tuning process. We call the tuned model in the experiment '25\% instruction data' as 'Llama-3-physician-8B instruct'. In the following paragraphs, we will compare it with other baselines. 

\paragraph{Baselines} For instruction-tuning, we consider instruction-tuned models like Mistral-7B-instruct~\citep{Jiang2023Mistral7}, Zephyr-7B-$\beta$-instruct~\citep{tunstall2023zephyr}, PMC-Llama-7B \citep{Wu2023PMCLlamaTB}, BioMedGPT-LM 7B \citep{zhang2023biomedgpt}, Medalpaca-13B \citep{han2023medalpaca}, AlpaCare-13B \citep{zhang2023alpacareinstructiontuned}, Me-LLaMA-13B chat\citep{xie2024me}, Llama-3-8B instruct~\citep{LLaMa-3}, and JSL-Med-Sft-Llama-3-8B~\citep{JSL-Med-Sft-Llama-3}. These LLMs are tuned with general instructions or medical task instructions. 
\begin{table}[h]
    \centering
    \resizebox{\textwidth}{!}{
        \begin{tabular}{lccccc}
            \toprule
            \textbf{Model} & \textbf{MMLU-Medical} & \textbf{PubMedQA} & \textbf{MedMCQA} & \textbf{MedQA-4-Option} & \textbf{Avg} \\
            \midrule
              Mistral-7B-instruct~\citep{Jiang2023Mistral7} & 55.8 & 17.8 & 40.2 & 41.1 & 37.5 \\
            Zephyr-7B-instruct-$\beta$~\citep{tunstall2023zephyr} & 63.3 & 46.0 & 43.0 & 48.5 & 48.7 \\
            PMC-Llama-7B~\citep{Wu2023PMCLlamaTB} & 59.7 & 59.2 & 57.6 & 49.2 & 53.6 \\
            Medalpaca-13B \citep{han2023medalpaca} & 55.2 &50.4&21.2&20.2&36.7\\ 
            AlpaCare-13B \citep{zhang2023alpacareinstructiontuned} & 60.2 &53.8&38.5&30.4&45.7\\
            BioMedGPT-LM 7B~\citep{zhang2023biomedgpt}&52.0&58.6&34.9&39.3&46.2\\
            Me-Llama-13B~\citep{xie2024me}&-&70.0&44.9&42.7&-\\
            Llama-3-8B instruct &82.0&74.6&57.1&60.3&68.5\\
            JSL-Med-Sft-Llama-3-8B~\citep{JSL-Med-Sft-Llama-3} &83.0&75.4&57.5&59.7&68.9 \\
            \midrule
            GPT-3.5-turbo-1106 &74.0 & 72.6&34.9&39.3& 60.6\\  
            GPT-4~\citep{openai2023gpt4} &\bf85.5 & 69.2&69.5&\bf83.9&\bf77.0 \\
            \midrule
            Llama-3-physician-8B instruct (ours) & 80.0 & \bf76.0 & \bf80.2 & 60.3 & 74.1 \\
            \bottomrule
        \end{tabular}
    }
    \caption{Accuracy comparison for question-answering tasks in the instruction-tuning setting.}
    \label{tab:instruction_tuned_acc}
    \vspace{-3mm}
\end{table}
\begin{table}[h]
    \centering
    \resizebox{\textwidth}{!}{
        \begin{tabular}{lcccc}
            \toprule
            
            \textbf{Task type} & \textbf{Classification} & \textbf{Relation extraction} & \textbf{Natural Language Inference} & \textbf{Summarization}  \\
            \midrule
            \textbf{Datasets} & \textbf{HOC} & \textbf{DDI-2013} & \textbf{BioNLI} & \textbf{MIMIC-CXR}  \\
            \midrule
              Mistral-7B-instruct~\citep{Jiang2023Mistral7} & 35.8 & 14.1 & 16.7 & 12.5  \\
            Zephyr-7B-instruct-$\beta$~\citep{tunstall2023zephyr} & 26.1 & 19.4 & 19.9 & 10.5  \\
            PMC-Llama-7B~\citep{Wu2023PMCLlamaTB}  & 18.4 & 14.7 & 15.9 & 13.9 \\
            Medalpaca-13B \citep{han2023medalpaca} & 24.6 &5.8&16.4&1.0\\ 
            AlpaCare-13B \citep{zhang2023alpacareinstructiontuned} & 26.7 &11.0&17.0&13.4\\
            BioMedGPT-LM 7B~\citep{zhang2023biomedgpt}&23.4&15.5&17.9&6.2\\
            Me-Llama-13B~\citep{xie2024me}&33.5&21.4&19.5&\bf40.0\\
            JSL-Med-Sft-Llama-3-8B~\citep{JSL-Med-Sft-Llama-3} &25.6&19.7&16.6&13.8 \\
            Llama-3-8B instruct &31.0&15.1&18.8&10.3\\
            \midrule
            GPT-3.5-turbo-1106 &54.5 & 21.6&31.7&13.5\\  
            GPT-4~\citep{openai2023gpt4} &60.2 & 29.2&57.8&15.2\\
            \midrule
            Llama-3-physician-8B instruct (ours) & \bf78.9& \bf33.6 & \bf76.2 & 37.7 \\
            \bottomrule
        \end{tabular}
    }
    \caption{Performance comparison for general medical tasks in the instruction-tuning setting. For BioNLI, DDI 2023, and HOC tasks, we report macro-F1. For MIMIC-CXR summarization tasks, we report Rouge-L as the result.}
    \label{tab:instruction_tuned_performance}
    \vspace{-3mm}
\end{table}
\paragraph{Results} 
We download the baselines' official models/deploy their APIs and then test their task performance using lm-eval-harnesses and Me-Llama's evaluation frameworks. If the paper does not release its model, we copy the results from the original paper (e.g., Me-Llama). From Table~\ref{tab:instruction_tuned_acc}, we find that our model outperforms other open-source baselines in question-answering tasks by a clear margin. Additionally, our model's average performance is close to that of GPT-4. Furthermore, in Table~\ref{tab:instruction_tuned_performance}, we observe that our model significantly outperforms GPT-4 in medical classification, relation extraction, natural language inference, and summarization tasks. This demonstrates the significant advantage of our model in processing diverse medical tasks. Finally, compared to these closed-source LLMs and larger open-source LLMs, our 8B model has the potential advantage of being deployed on users' local devices, reducing the risk of leaking personal healthcare information.


\section{Conclusion}
Our paper explores the behavior of LLMs when continually pre-training them on a new domain's corpus and observes the stability gap, a phenomenon marked by a significant initial performance drop followed by a slow recovery. We explain it from the view of plasticity and stability gradients and then propose three strategies that effectively improve the LLM's domain performance and reduce computational costs by overcoming the stability gap. Furthermore, we deploy our strategies on the newest Llama-3-8B model, which achieves the strongest performance among open-source baselines of similar model scales and outperforms the closed-source GPT-3.5 model. 

\paragraph{Limitations and Potential impacts}
Ideally, knowing the pre-training data mixture could maximize the outcome of our method, but most strong open-source LLMs didn't provide their training data mixture.
Our Llama-3-8B experiment shows we can  still improve significantly in this scenario.
Due to limitations in computing resources, we plan to verify our conclusions and strategies on larger LLMs in the future. 
Our strategies are designed to address the machine learning problem of the stability gap, and we do not see any potential risks. 
The datasets and base models used in this paper will be open-sourced. 
Although we do not consider our model to be ready for real-world medical use in its current form, we are releasing it to the research community to promote work on large language models for the medical domain and the safety of language models in medical applications. 

\bibliographystyle{plainnat}
\bibliography{ref}

\begin{thebibliography}{96}
\providecommand{\natexlab}[1]{#1}
\providecommand{\url}[1]{\texttt{#1}}
\expandafter\ifx\csname urlstyle\endcsname\relax
  \providecommand{\doi}[1]{doi: #1}\else
  \providecommand{\doi}{doi: \begingroup \urlstyle{rm}\Url}\fi

\bibitem[Baker et~al.(2016)Baker, Silins, Guo, Ali, H{\"o}gberg, Stenius, and Korhonen]{baker2016automatic}
Simon Baker, Ilona Silins, Yufan Guo, Imran Ali, Johan H{\"o}gberg, Ulla Stenius, and Anna Korhonen.
\newblock Automatic semantic classification of scientific literature according to the hallmarks of cancer.
\newblock \emph{Bioinformatics}, 32\penalty0 (3):\penalty0 432--440, 2016.

\bibitem[Bastan et~al.(2022)Bastan, Surdeanu, and Balasubramanian]{bastan2022bionli}
Mohaddeseh Bastan, Mihai Surdeanu, and Niranjan Balasubramanian.
\newblock Bionli: Generating a biomedical nli dataset using lexico-semantic constraints for adversarial examples.
\newblock \emph{arXiv preprint arXiv:2210.14814}, 2022.

\bibitem[Biderman et~al.(2023{\natexlab{a}})Biderman, Schoelkopf, Anthony, Bradley, O'Brien, Hallahan, Khan, Purohit, Prashanth, Raff, Skowron, Sutawika, and van~der Wal]{Biderman2023PythiaAS}
Stella Biderman, Hailey Schoelkopf, Quentin~G. Anthony, Herbie Bradley, Kyle O'Brien, Eric Hallahan, Mohammad~Aflah Khan, Shivanshu Purohit, USVSN~Sai Prashanth, Edward Raff, Aviya Skowron, Lintang Sutawika, and Oskar van~der Wal.
\newblock Pythia: A suite for analyzing large language models across training and scaling.
\newblock \emph{ArXiv}, abs/2304.01373, 2023{\natexlab{a}}.
\newblock URL \url{https://api.semanticscholar.org/CorpusID:257921893}.

\bibitem[Biderman et~al.(2023{\natexlab{b}})Biderman, Schoelkopf, Anthony, Bradley, O’Brien, Hallahan, Khan, Purohit, Prashanth, Raff, et~al.]{biderman2023pythia}
Stella Biderman, Hailey Schoelkopf, Quentin~Gregory Anthony, Herbie Bradley, Kyle O’Brien, Eric Hallahan, Mohammad~Aflah Khan, Shivanshu Purohit, USVSN~Sai Prashanth, Edward Raff, et~al.
\newblock Pythia: A suite for analyzing large language models across training and scaling.
\newblock In \emph{International Conference on Machine Learning}, pages 2397--2430. PMLR, 2023{\natexlab{b}}.

\bibitem[Bisk et~al.(2020)Bisk, Zellers, Gao, Choi, et~al.]{bisk2020piqa}
Yonatan Bisk, Rowan Zellers, Jianfeng Gao, Yejin Choi, et~al.
\newblock Piqa: Reasoning about physical commonsense in natural language.
\newblock In \emph{Proceedings of the AAAI conference on artificial intelligence}, volume~34, pages 7432--7439, 2020.

\bibitem[Buzzega et~al.(2020)Buzzega, Boschini, Porrello, Abati, and Calderara]{buzzega2020dark}
Pietro Buzzega, Matteo Boschini, Angelo Porrello, Davide Abati, and Simone Calderara.
\newblock Dark experience for general continual learning: a strong, simple baseline.
\newblock \emph{Advances in neural information processing systems}, 33:\penalty0 15920--15930, 2020.

\bibitem[Buzzega et~al.(2021)Buzzega, Boschini, Porrello, and Calderara]{buzzega2021rethinking}
Pietro Buzzega, Matteo Boschini, Angelo Porrello, and Simone Calderara.
\newblock Rethinking experience replay: a bag of tricks for continual learning.
\newblock In \emph{2020 25th International Conference on Pattern Recognition (ICPR)}, pages 2180--2187. IEEE, 2021.

\bibitem[Caccia et~al.(2021)Caccia, Aljundi, Asadi, Tuytelaars, Pineau, and Belilovsky]{caccia2021new}
Lucas Caccia, Rahaf Aljundi, Nader Asadi, Tinne Tuytelaars, Joelle Pineau, and Eugene Belilovsky.
\newblock New insights on reducing abrupt representation change in online continual learning.
\newblock \emph{arXiv preprint arXiv:2104.05025}, 2021.

\bibitem[Chen et~al.(2023{\natexlab{a}})Chen, Zhou, Du, Huang, Laudon, Chen, and Cui]{chen2023lifelong}
Wuyang Chen, Yanqi Zhou, Nan Du, Yanping Huang, James Laudon, Zhifeng Chen, and Claire Cui.
\newblock Lifelong language pretraining with distribution-specialized experts.
\newblock In \emph{International Conference on Machine Learning}, pages 5383--5395. PMLR, 2023{\natexlab{a}}.

\bibitem[Chen et~al.(2024)Chen, Hu, and Zhang]{Chen2024CompressingLL}
Xiaodong Chen, Yuxuan Hu, and Jing Zhang.
\newblock Compressing large language models by streamlining the unimportant layer.
\newblock \emph{ArXiv}, abs/2403.19135, 2024.
\newblock URL \url{https://api.semanticscholar.org/CorpusID:268733054}.

\bibitem[Chen et~al.(2023{\natexlab{b}})Chen, Cano, Romanou, Bonnet, Matoba, Salvi, Pagliardini, Fan, K{\"o}pf, Mohtashami, et~al.]{chen2023meditron}
Zeming Chen, Alejandro~Hern{\'a}ndez Cano, Angelika Romanou, Antoine Bonnet, Kyle Matoba, Francesco Salvi, Matteo Pagliardini, Simin Fan, Andreas K{\"o}pf, Amirkeivan Mohtashami, et~al.
\newblock Meditron-70b: Scaling medical pretraining for large language models.
\newblock \emph{arXiv preprint arXiv:2311.16079}, 2023{\natexlab{b}}.

\bibitem[Chen et~al.(2023{\natexlab{c}})Chen, Hernández-Cano, Romanou, Bonnet, Matoba, Salvi, Pagliardini, Fan, Köpf, Mohtashami, Sallinen, Sakhaeirad, Swamy, Krawczuk, Bayazit, Marmet, Montariol, Hartley, Jaggi, and Bosselut]{chen2023meditron70b}
Zeming Chen, Alejandro Hernández-Cano, Angelika Romanou, Antoine Bonnet, Kyle Matoba, Francesco Salvi, Matteo Pagliardini, Simin Fan, Andreas Köpf, Amirkeivan Mohtashami, Alexandre Sallinen, Alireza Sakhaeirad, Vinitra Swamy, Igor Krawczuk, Deniz Bayazit, Axel Marmet, Syrielle Montariol, Mary-Anne Hartley, Martin Jaggi, and Antoine Bosselut.
\newblock Meditron-70b: Scaling medical pretraining for large language models, 2023{\natexlab{c}}.

\bibitem[Clark et~al.(2019)Clark, Lee, Chang, Kwiatkowski, Collins, and Toutanova]{clark2019boolq}
Christopher Clark, Kenton Lee, Ming-Wei Chang, Tom Kwiatkowski, Michael Collins, and Kristina Toutanova.
\newblock Boolq: Exploring the surprising difficulty of natural yes/no questions.
\newblock In \emph{NAACL}, 2019.

\bibitem[Clark et~al.(2018)Clark, Cowhey, Etzioni, Khot, Sabharwal, Schoenick, and Tafjord]{clark2018think}
Peter Clark, Isaac Cowhey, Oren Etzioni, Tushar Khot, Ashish Sabharwal, Carissa Schoenick, and Oyvind Tafjord.
\newblock Think you have solved question answering? try arc, the ai2 reasoning challenge.
\newblock \emph{arXiv preprint arXiv:1803.05457}, 2018.

\bibitem[Dao(2023)]{dao2023flashattention}
Tri Dao.
\newblock Flashattention-2: Faster attention with better parallelism and work partitioning.
\newblock \emph{arXiv preprint arXiv:2307.08691}, 2023.

\bibitem[De~Lange et~al.(2022)De~Lange, van~de Ven, and Tuytelaars]{de2022continual}
Matthias De~Lange, Gido van~de Ven, and Tinne Tuytelaars.
\newblock Continual evaluation for lifelong learning: Identifying the stability gap.
\newblock \emph{arXiv preprint arXiv:2205.13452}, 2022.

\bibitem[Devlin et~al.(2019)Devlin, Chang, Lee, and Toutanova]{Devlin2019BERTPO}
Jacob Devlin, Ming-Wei Chang, Kenton Lee, and Kristina Toutanova.
\newblock Bert: Pre-training of deep bidirectional transformers for language understanding.
\newblock In \emph{North American Chapter of the Association for Computational Linguistics}, 2019.
\newblock URL \url{https://api.semanticscholar.org/CorpusID:52967399}.

\bibitem[Durbin(2024)]{airoboros}
Jon Durbin.
\newblock airoboros: Customizable implementation of the self-instruct paper., 2024.
\newblock URL \url{https://huggingface.co/datasets/jondurbin/airoboros-3.2}.

\bibitem[Fu et~al.(2024)Fu, Panda, Niu, Yue, Hajishirzi, Kim, and Peng]{fu2024data}
Yao Fu, Rameswar Panda, Xinyao Niu, Xiang Yue, Hannaneh Hajishirzi, Yoon Kim, and Hao Peng.
\newblock Data engineering for scaling language models to 128k context.
\newblock \emph{arXiv preprint arXiv:2402.10171}, 2024.

\bibitem[Gao et~al.(2020)Gao, Biderman, Black, Golding, Hoppe, Foster, Phang, He, Thite, Nabeshima, Presser, and Leahy]{Gao2020ThePA}
Leo Gao, Stella Biderman, Sid Black, Laurence Golding, Travis Hoppe, Charles Foster, Jason Phang, Horace He, Anish Thite, Noa Nabeshima, Shawn Presser, and Connor Leahy.
\newblock The pile: An 800gb dataset of diverse text for language modeling.
\newblock \emph{ArXiv}, abs/2101.00027, 2020.
\newblock URL \url{https://api.semanticscholar.org/CorpusID:230435736}.

\bibitem[Gao et~al.(2023)Gao, Tow, Abbasi, Biderman, Black, DiPofi, Foster, Golding, Hsu, Le~Noac'h, Li, McDonell, Muennighoff, Ociepa, Phang, Reynolds, Schoelkopf, Skowron, Sutawika, Tang, Thite, Wang, Wang, and Zou]{eval-harness}
Leo Gao, Jonathan Tow, Baber Abbasi, Stella Biderman, Sid Black, Anthony DiPofi, Charles Foster, Laurence Golding, Jeffrey Hsu, Alain Le~Noac'h, Haonan Li, Kyle McDonell, Niklas Muennighoff, Chris Ociepa, Jason Phang, Laria Reynolds, Hailey Schoelkopf, Aviya Skowron, Lintang Sutawika, Eric Tang, Anish Thite, Ben Wang, Kevin Wang, and Andy Zou.
\newblock A framework for few-shot language model evaluation, 12 2023.
\newblock URL \url{https://zenodo.org/records/10256836}.

\bibitem[Geng and Liu(2023)]{openlm2023openLlama}
Xinyang Geng and Hao Liu.
\newblock Openllama: An open reproduction of llama, May 2023.
\newblock URL \url{https://github.com/openlm-research/open_llama}.

\bibitem[Guo et~al.(2022)Guo, Liu, and Zhao]{guo2022online}
Yiduo Guo, Bing Liu, and Dongyan Zhao.
\newblock Online continual learning through mutual information maximization.
\newblock In \emph{International Conference on Machine Learning}, pages 8109--8126. PMLR, 2022.

\bibitem[Guo et~al.(2023)Guo, Zhang, Liang, Zhao, and Nan]{guo2023pptc}
Yiduo Guo, Zekai Zhang, Yaobo Liang, Dongyan Zhao, and Duan Nan.
\newblock Pptc benchmark: Evaluating large language models for powerpoint task completion.
\newblock \emph{arXiv preprint arXiv:2311.01767}, 2023.

\bibitem[Gupta et~al.(2023)Gupta, Th'erien, Ibrahim, Richter, Anthony, Belilovsky, Rish, and Lesort]{Gupta2023ContinualPO}
Kshitij Gupta, Benjamin Th'erien, Adam Ibrahim, Mats~L. Richter, Quentin~G. Anthony, Eugene Belilovsky, Irina Rish, and Timoth{\'e}e Lesort.
\newblock Continual pre-training of large language models: How to (re)warm your model?
\newblock \emph{ArXiv}, abs/2308.04014, 2023.
\newblock URL \url{https://api.semanticscholar.org/CorpusID:260704601}.

\bibitem[Gururajan et~al.(2024)Gururajan, Lopez-Cuena, Bayarri-Planas, Tormos, Hinjos, Bernabeu-Perez, Arias-Duart, Martin-Torres, Urcelay-Ganzabal, Gonzalez-Mallo, et~al.]{gururajan2024aloe}
Ashwin~Kumar Gururajan, Enrique Lopez-Cuena, Jordi Bayarri-Planas, Adrian Tormos, Daniel Hinjos, Pablo Bernabeu-Perez, Anna Arias-Duart, Pablo~Agustin Martin-Torres, Lucia Urcelay-Ganzabal, Marta Gonzalez-Mallo, et~al.
\newblock Aloe: A family of fine-tuned open healthcare llms.
\newblock \emph{arXiv preprint arXiv:2405.01886}, 2024.

\bibitem[Han et~al.(2023)Han, Adams, Papaioannou, Grundmann, Oberhauser, L{\"o}ser, Truhn, and Bressem]{han2023medalpaca}
Tianyu Han, Lisa~C Adams, Jens-Michalis Papaioannou, Paul Grundmann, Tom Oberhauser, Alexander L{\"o}ser, Daniel Truhn, and Keno~K Bressem.
\newblock Medalpaca--an open-source collection of medical conversational ai models and training data.
\newblock \emph{arXiv preprint arXiv:2304.08247}, 2023.

\bibitem[Harun and Kanan(2023)]{harun2023overcoming}
Md~Yousuf Harun and Christopher Kanan.
\newblock Overcoming the stability gap in continual learning.
\newblock \emph{arXiv preprint arXiv:2306.01904}, 2023.

\bibitem[Heafield(2011)]{Heafield2011KenLMFA}
Kenneth Heafield.
\newblock Kenlm: Faster and smaller language model queries.
\newblock In \emph{WMT@EMNLP}, 2011.
\newblock URL \url{https://api.semanticscholar.org/CorpusID:8313873}.

\bibitem[Hendrycks et~al.(2020{\natexlab{a}})Hendrycks, Burns, Basart, Zou, Mazeika, Song, and Steinhardt]{hendrycks2020measuring}
Dan Hendrycks, Collin Burns, Steven Basart, Andy Zou, Mantas Mazeika, Dawn Song, and Jacob Steinhardt.
\newblock Measuring massive multitask language understanding.
\newblock \emph{arXiv preprint arXiv:2009.03300}, 2020{\natexlab{a}}.

\bibitem[Hendrycks et~al.(2020{\natexlab{b}})Hendrycks, Burns, Basart, Zou, Mazeika, Song, and Steinhardt]{Hendrycks2020MeasuringMM}
Dan Hendrycks, Collin Burns, Steven Basart, Andy Zou, Mantas Mazeika, Dawn~Xiaodong Song, and Jacob Steinhardt.
\newblock Measuring massive multitask language understanding.
\newblock \emph{ArXiv}, abs/2009.03300, 2020{\natexlab{b}}.
\newblock URL \url{https://api.semanticscholar.org/CorpusID:221516475}.

\bibitem[Hernandez et~al.(2022)Hernandez, Brown, Conerly, DasSarma, Drain, El-Showk, Elhage, Hatfield-Dodds, Henighan, Hume, et~al.]{hernandez2022scaling}
Danny Hernandez, Tom Brown, Tom Conerly, Nova DasSarma, Dawn Drain, Sheer El-Showk, Nelson Elhage, Zac Hatfield-Dodds, Tom Henighan, Tristan Hume, et~al.
\newblock Scaling laws and interpretability of learning from repeated data.
\newblock \emph{arXiv preprint arXiv:2205.10487}, 2022.

\bibitem[Hewitt and Manning(2019)]{Hewitt2019ASP}
John Hewitt and Christopher~D. Manning.
\newblock A structural probe for finding syntax in word representations.
\newblock In \emph{North American Chapter of the Association for Computational Linguistics}, 2019.
\newblock URL \url{https://api.semanticscholar.org/CorpusID:106402715}.

\bibitem[Hoffmann et~al.(2022)Hoffmann, Borgeaud, Mensch, Buchatskaya, Cai, Rutherford, de~Las~Casas, Hendricks, Welbl, Clark, Hennigan, Noland, Millican, van~den Driessche, Damoc, Guy, Osindero, Simonyan, Elsen, Rae, Vinyals, and Sifre]{Hoffmann2022TrainingCL}
Jordan Hoffmann, Sebastian Borgeaud, Arthur Mensch, Elena Buchatskaya, Trevor Cai, Eliza Rutherford, Diego de~Las~Casas, Lisa~Anne Hendricks, Johannes Welbl, Aidan Clark, Tom Hennigan, Eric Noland, Katie Millican, George van~den Driessche, Bogdan Damoc, Aurelia Guy, Simon Osindero, Karen Simonyan, Erich Elsen, Jack~W. Rae, Oriol Vinyals, and L.~Sifre.
\newblock Training compute-optimal large language models.
\newblock \emph{ArXiv}, abs/2203.15556, 2022.
\newblock URL \url{https://api.semanticscholar.org/CorpusID:247778764}.

\bibitem[Huang et~al.(2023)Huang, Tao, An, Zhang, Jiang, Chen, Wu, and Feng]{huang2023lawyer}
Quzhe Huang, Mingxu Tao, Zhenwei An, Chen Zhang, Cong Jiang, Zhibin Chen, Zirui Wu, and Yansong Feng.
\newblock Lawyer llama technical report.
\newblock \emph{arXiv preprint arXiv:2305.15062}, 2023.

\bibitem[Ibrahim et~al.(2024)Ibrahim, Th{\'e}rien, Gupta, Richter, Anthony, Lesort, Belilovsky, and Rish]{ibrahim2024simple}
Adam Ibrahim, Benjamin Th{\'e}rien, Kshitij Gupta, Mats~L Richter, Quentin Anthony, Timoth{\'e}e Lesort, Eugene Belilovsky, and Irina Rish.
\newblock Simple and scalable strategies to continually pre-train large language models.
\newblock \emph{arXiv preprint arXiv:2403.08763}, 2024.

\bibitem[Jiang et~al.(2023)Jiang, Sablayrolles, Mensch, Bamford, Chaplot, de~Las~Casas, Bressand, Lengyel, Lample, Saulnier, Lavaud, Lachaux, Stock, Scao, Lavril, Wang, Lacroix, and Sayed]{Jiang2023Mistral7}
Albert~Qiaochu Jiang, Alexandre Sablayrolles, Arthur Mensch, Chris Bamford, Devendra~Singh Chaplot, Diego de~Las~Casas, Florian Bressand, Gianna Lengyel, Guillaume Lample, Lucile Saulnier, L'elio~Renard Lavaud, Marie-Anne Lachaux, Pierre Stock, Teven~Le Scao, Thibaut Lavril, Thomas Wang, Timoth{\'e}e Lacroix, and William~El Sayed.
\newblock Mistral 7b.
\newblock \emph{ArXiv}, abs/2310.06825, 2023.
\newblock URL \url{https://api.semanticscholar.org/CorpusID:263830494}.

\bibitem[Jiang et~al.(2024)Jiang, Sun, Shi, Rodriguez, Zhou, Neubig, Lin, Yih, and Iyer]{jiang2024instruction}
Zhengbao Jiang, Zhiqing Sun, Weijia Shi, Pedro Rodriguez, Chunting Zhou, Graham Neubig, Xi~Victoria Lin, Wen-tau Yih, and Srinivasan Iyer.
\newblock Instruction-tuned language models are better knowledge learners.
\newblock \emph{arXiv preprint arXiv:2402.12847}, 2024.

\bibitem[Jin et~al.(2021{\natexlab{a}})Jin, Pan, Oufattole, Weng, Fang, and Szolovits]{jin2021disease}
Di~Jin, Eileen Pan, Nassim Oufattole, Wei-Hung Weng, Hanyi Fang, and Peter Szolovits.
\newblock What disease does this patient have? a large-scale open domain question answering dataset from medical exams.
\newblock \emph{Applied Sciences}, 11\penalty0 (14):\penalty0 6421, 2021{\natexlab{a}}.

\bibitem[Jin et~al.(2019)Jin, Dhingra, Liu, Cohen, and Lu]{jin2019pubmedqa}
Qiao Jin, Bhuwan Dhingra, Zhengping Liu, William~W Cohen, and Xinghua Lu.
\newblock Pubmedqa: A dataset for biomedical research question answering.
\newblock \emph{arXiv preprint arXiv:1909.06146}, 2019.

\bibitem[Jin et~al.(2021{\natexlab{b}})Jin, Zhang, Zhu, Xiao, Li, Wei, Arnold, and Ren]{jin2021lifelong}
Xisen Jin, Dejiao Zhang, Henghui Zhu, Wei Xiao, Shang-Wen Li, Xiaokai Wei, Andrew Arnold, and Xiang Ren.
\newblock Lifelong pretraining: Continually adapting language models to emerging corpora.
\newblock \emph{arXiv preprint arXiv:2110.08534}, 2021{\natexlab{b}}.

\bibitem[johnsnowlabs(2024)]{JSL-Med-Sft-Llama-3}
johnsnowlabs.
\newblock Jsl-med-sft-llama-3, a finetuned medical llm developed by john snow labs, 2024.
\newblock URL \url{https://huggingface.co/johnsnowlabs/JSL-Med-Sft-Llama-3-8B}.

\bibitem[Johnson et~al.(2019)Johnson, Pollard, Berkowitz, Greenbaum, Lungren, Deng, Mark, and Horng]{johnson2019mimic}
Alistair~EW Johnson, Tom~J Pollard, Seth~J Berkowitz, Nathaniel~R Greenbaum, Matthew~P Lungren, Chih-ying Deng, Roger~G Mark, and Steven Horng.
\newblock Mimic-cxr, a de-identified publicly available database of chest radiographs with free-text reports.
\newblock \emph{Scientific data}, 6\penalty0 (1):\penalty0 317, 2019.

\bibitem[Kaplan et~al.(2020)Kaplan, McCandlish, Henighan, Brown, Chess, Child, Gray, Radford, Wu, and Amodei]{Kaplan2020ScalingLF}
Jared Kaplan, Sam McCandlish, T.~J. Henighan, Tom~B. Brown, Benjamin Chess, Rewon Child, Scott Gray, Alec Radford, Jeff Wu, and Dario Amodei.
\newblock Scaling laws for neural language models.
\newblock \emph{ArXiv}, abs/2001.08361, 2020.
\newblock URL \url{https://api.semanticscholar.org/CorpusID:210861095}.

\bibitem[Ke et~al.(2022)Ke, Lin, Shao, Xu, Shu, and Liu]{ke2022continual}
Zixuan Ke, Haowei Lin, Yijia Shao, Hu~Xu, Lei Shu, and Bing Liu.
\newblock Continual training of language models for few-shot learning.
\newblock In \emph{Empirical Methods in Natural Language Processing (EMNLP)}, 2022.

\bibitem[Ke et~al.(2023)Ke, Shao, Lin, Xu, Shu, and Liu]{ke2023adapting}
Zixuan Ke, Yijia Shao, Haowei Lin, Hu~Xu, Lei Shu, and Bing Liu.
\newblock Adapting a language model while preserving its general knowledge.
\newblock \emph{arXiv preprint arXiv:2301.08986}, 2023.

\bibitem[Kirkpatrick et~al.(2017)Kirkpatrick, Pascanu, Rabinowitz, Veness, Desjardins, Rusu, Milan, Quan, Ramalho, Grabska-Barwinska, et~al.]{kirkpatrick2017overcoming}
James Kirkpatrick, Razvan Pascanu, Neil Rabinowitz, Joel Veness, Guillaume Desjardins, Andrei~A Rusu, Kieran Milan, John Quan, Tiago Ramalho, Agnieszka Grabska-Barwinska, et~al.
\newblock Overcoming catastrophic forgetting in neural networks.
\newblock \emph{Proceedings of the national academy of sciences}, 114\penalty0 (13):\penalty0 3521--3526, 2017.

\bibitem[Labrak et~al.(2024)Labrak, Bazoge, Morin, Gourraud, Rouvier, and Dufour]{labrak2024biomistral}
Yanis Labrak, Adrien Bazoge, Emmanuel Morin, Pierre-Antoine Gourraud, Mickael Rouvier, and Richard Dufour.
\newblock Biomistral: A collection of open-source pretrained large language models for medical domains.
\newblock \emph{arXiv preprint arXiv:2402.10373}, 2024.

\bibitem[Lai et~al.(2017)Lai, Xie, Liu, Yang, and Hovy]{lai2017large}
Guokun Lai, Qizhe Xie, Hanxiao Liu, Yiming Yang, and Eduard Hovy.
\newblock Race: Large-scale reading comprehension dataset from examinations.
\newblock \emph{arXiv preprint arXiv:1704.04683}, 2017.

\bibitem[Lange et~al.(2022)Lange, van~de Ven, and Tuytelaars]{DeLange2022ContinualEF}
Matthias~De Lange, Gido~M. van~de Ven, and Tinne Tuytelaars.
\newblock Continual evaluation for lifelong learning: Identifying the stability gap.
\newblock \emph{ArXiv}, abs/2205.13452, 2022.
\newblock URL \url{https://api.semanticscholar.org/CorpusID:249097739}.

\bibitem[Lin et~al.(2024)Lin, Gou, Gong, Liu, Shen, Xu, Lin, Yang, Jiao, Duan, et~al.]{lin2024rho}
Zhenghao Lin, Zhibin Gou, Yeyun Gong, Xiao Liu, Yelong Shen, Ruochen Xu, Chen Lin, Yujiu Yang, Jian Jiao, Nan Duan, et~al.
\newblock Rho-1: Not all tokens are what you need.
\newblock \emph{arXiv preprint arXiv:2404.07965}, 2024.

\bibitem[Ling et~al.(2023)Ling, Zhao, Lu, Deng, Zheng, Wang, Chowdhury, Li, Cui, Zhang, yu~Zhao, Panalkar, Cheng, Wang, Liu, Chen, Chen, White, Gu, Pei, Yang, and Zhao]{Ling2023DomainSA}
Chen Ling, Xujiang Zhao, Jiaying Lu, Chengyuan Deng, Can Zheng, Junxiang Wang, Tanmoy Chowdhury, Yun-Qing Li, Hejie Cui, Xuchao Zhang, Tian yu~Zhao, Amit Panalkar, Wei Cheng, Haoyu Wang, Yanchi Liu, Zhengzhang Chen, Haifeng Chen, Chris White, Quanquan Gu, Jian Pei, Carl Yang, and Liang Zhao.
\newblock Domain specialization as the key to make large language models disruptive: A comprehensive survey.
\newblock 2023.
\newblock URL \url{https://api.semanticscholar.org/CorpusID:259502302}.

\bibitem[Liu et~al.(2024)Liu, Zeng, He, Jiang, and He]{liu2024what}
Wei Liu, Weihao Zeng, Keqing He, Yong Jiang, and Junxian He.
\newblock What makes good data for alignment? a comprehensive study of automatic data selection in instruction tuning.
\newblock In \emph{The Twelfth International Conference on Learning Representations}, 2024.
\newblock URL \url{https://openreview.net/forum?id=BTKAeLqLMw}.

\bibitem[Liu et~al.(2023)Liu, Yu, Zhang, Xu, Lei, Lai, Gu, Ding, Men, Yang, Zhang, Deng, Zeng, Du, Zhang, Shen, Zhang, Su, Sun, Huang, Dong, and Tang]{liu2023agentbench}
Xiao Liu, Hao Yu, Hanchen Zhang, Yifan Xu, Xuanyu Lei, Hanyu Lai, Yu~Gu, Hangliang Ding, Kaiwen Men, Kejuan Yang, Shudan Zhang, Xiang Deng, Aohan Zeng, Zhengxiao Du, Chenhui Zhang, Sheng Shen, Tianjun Zhang, Yu~Su, Huan Sun, Minlie Huang, Yuxiao Dong, and Jie Tang.
\newblock Agentbench: Evaluating llms as agents.
\newblock \emph{arXiv preprint arXiv: 2308.03688}, 2023.

\bibitem[Meta(2024)]{LLaMa-3}
Meta.
\newblock Introducing meta llama 3: The most capable openly available llm to date, 2024.
\newblock URL \url{https://ai.meta.com/blog/meta-llama-3/}.

\bibitem[Mihaylov et~al.(2018)Mihaylov, Clark, Khot, and Sabharwal]{Mihaylov2018CanAS}
Todor Mihaylov, Peter Clark, Tushar Khot, and Ashish Sabharwal.
\newblock Can a suit of armor conduct electricity? a new dataset for open book question answering.
\newblock In \emph{Conference on Empirical Methods in Natural Language Processing}, 2018.
\newblock URL \url{https://api.semanticscholar.org/CorpusID:52183757}.

\bibitem[Muennighoff et~al.(2024)Muennighoff, Rush, Barak, Le~Scao, Tazi, Piktus, Pyysalo, Wolf, and Raffel]{muennighoff2024scaling}
Niklas Muennighoff, Alexander Rush, Boaz Barak, Teven Le~Scao, Nouamane Tazi, Aleksandra Piktus, Sampo Pyysalo, Thomas Wolf, and Colin~A Raffel.
\newblock Scaling data-constrained language models.
\newblock \emph{Advances in Neural Information Processing Systems}, 36, 2024.

\bibitem[Nguyen et~al.(2023)Nguyen, Ting, Ciucă, O'Neill, Sun, Jablonska, Kruk, Perkowski, Miller, Li, Peek, Iyer, R'o.za'nski, Khetarpal, Zaman, Brodrick, M'endez, Bui, Goodman, Accomazzi, Naiman, Cranney, Schawinski, and UniverseTBD]{Nguyen2023AstroLlamaTS}
Tuan~Dung Nguyen, Yuan-Sen Ting, Ioana Ciucă, Charlie O'Neill, Ze-Chang Sun, Maja Jablonska, Sandor~J. Kruk, Ernest Perkowski, Jack~W. Miller, Jason Li, Josh Peek, Kartheik~G. Iyer, Tomasz R'o.za'nski, Pranav Khetarpal, Sharaf Zaman, David Brodrick, Sergio J.~Rodr'iguez M'endez, Thang Bui, Alyssa Goodman, Alberto Accomazzi, Jill~P. Naiman, Jesse Cranney, Kevin Schawinski, and UniverseTBD.
\newblock Astrollama: Towards specialized foundation models in astronomy.
\newblock \emph{ArXiv}, abs/2309.06126, 2023.
\newblock URL \url{https://api.semanticscholar.org/CorpusID:261696577}.

\bibitem[OpenAI(2023)]{openai2023gpt4}
OpenAI.
\newblock Gpt-4 technical report.
\newblock \emph{arXiv preprint arXiv:2303.08774}, 2023.

\bibitem[Pal et~al.(2022)Pal, Umapathi, and Sankarasubbu]{pmlr-v174-pal22a}
Ankit Pal, Logesh~Kumar Umapathi, and Malaikannan Sankarasubbu.
\newblock Medmcqa: A large-scale multi-subject multi-choice dataset for medical domain question answering.
\newblock In Gerardo Flores, George~H Chen, Tom Pollard, Joyce~C Ho, and Tristan Naumann, editors, \emph{Proceedings of the Conference on Health, Inference, and Learning}, volume 174 of \emph{Proceedings of Machine Learning Research}, pages 248--260. PMLR, 07--08 Apr 2022.
\newblock URL \url{https://proceedings.mlr.press/v174/pal22a.html}.

\bibitem[Penedo et~al.(2023)Penedo, Malartic, Hesslow, Cojocaru, Cappelli, Alobeidli, Pannier, Almazrouei, and Launay]{penedo2023refinedweb}
Guilherme Penedo, Quentin Malartic, Daniel Hesslow, Ruxandra Cojocaru, Alessandro Cappelli, Hamza Alobeidli, Baptiste Pannier, Ebtesam Almazrouei, and Julien Launay.
\newblock The refinedweb dataset for falcon llm: outperforming curated corpora with web data, and web data only.
\newblock \emph{arXiv preprint arXiv:2306.01116}, 2023.

\bibitem[Prabhu et~al.(2020)Prabhu, Torr, and Dokania]{prabhu2020gdumb}
Ameya Prabhu, Philip~HS Torr, and Puneet~K Dokania.
\newblock Gdumb: A simple approach that questions our progress in continual learning.
\newblock In \emph{Computer Vision--ECCV 2020: 16th European Conference, Glasgow, UK, August 23--28, 2020, Proceedings, Part II 16}, pages 524--540. Springer, 2020.

\bibitem[Qin et~al.(2023)Qin, Zhang, Zhang, Chen, Yasunaga, and Yang]{qin2023chatgpt}
Chengwei Qin, Aston Zhang, Zhuosheng Zhang, Jiaao Chen, Michihiro Yasunaga, and Diyi Yang.
\newblock Is chatgpt a general-purpose natural language processing task solver?
\newblock \emph{arXiv preprint arXiv:2302.06476}, 2023.

\bibitem[Roemmele et~al.(2011)Roemmele, Bejan, and Gordon]{roemmele2011choice}
Melissa Roemmele, Cosmin~Adrian Bejan, and Andrew~S Gordon.
\newblock Choice of plausible alternatives: An evaluation of commonsense causal reasoning.
\newblock In \emph{2011 AAAI Spring Symposium Series}, 2011.

\bibitem[Rolnick et~al.(2019)Rolnick, Ahuja, Schwarz, Lillicrap, and Wayne]{rolnick2019experience}
David Rolnick, Arun Ahuja, Jonathan Schwarz, Timothy Lillicrap, and Gregory Wayne.
\newblock Experience replay for continual learning.
\newblock \emph{Advances in neural information processing systems}, 32, 2019.

\bibitem[Sakaguchi et~al.(2021)Sakaguchi, Bras, Bhagavatula, and Choi]{ai2:winogrande}
Keisuke Sakaguchi, Ronan~Le Bras, Chandra Bhagavatula, and Yejin Choi.
\newblock Winogrande: An adversarial winograd schema challenge at scale.
\newblock \emph{Communications of the ACM}, 64\penalty0 (9):\penalty0 99--106, 2021.

\bibitem[Segura-Bedmar et~al.(2013)Segura-Bedmar, Mart{\'\i}nez~Fern{\'a}ndez, and Herrero~Zazo]{segura2013semeval}
Isabel Segura-Bedmar, Paloma Mart{\'\i}nez~Fern{\'a}ndez, and Mar{\'\i}a Herrero~Zazo.
\newblock Semeval-2013 task 9: Extraction of drug-drug interactions from biomedical texts (ddiextraction 2013).
\newblock Association for Computational Linguistics, 2013.

\bibitem[Shen et~al.(2023)Shen, Tao, Ma, Neiswanger, Hestness, Vassilieva, Soboleva, and Xing]{shen2023slimpajama}
Zhiqiang Shen, Tianhua Tao, Liqun Ma, Willie Neiswanger, Joel Hestness, Natalia Vassilieva, Daria Soboleva, and Eric Xing.
\newblock Slimpajama-dc: Understanding data combinations for llm training.
\newblock \emph{arXiv preprint arXiv:2309.10818}, 2023.

\bibitem[Shi et~al.(2024)Shi, Xu, Zhuang, Yu, Wu, Yang, and Wang]{Shi2024MedAdapterET}
Wenqi Shi, Ran Xu, Yuchen Zhuang, Yue Yu, Hang Wu, Carl Yang, and May~Dongmei Wang.
\newblock Medadapter: Efficient test-time adaptation of large language models towards medical reasoning.
\newblock 2024.
\newblock URL \url{https://api.semanticscholar.org/CorpusID:269605605}.

\bibitem[Sorscher et~al.(2022)Sorscher, Geirhos, Shekhar, Ganguli, and Morcos]{Sorscher2022BeyondNS}
Ben Sorscher, Robert Geirhos, Shashank Shekhar, Surya Ganguli, and Ari~S. Morcos.
\newblock Beyond neural scaling laws: beating power law scaling via data pruning.
\newblock \emph{ArXiv}, abs/2206.14486, 2022.
\newblock URL \url{https://api.semanticscholar.org/CorpusID:250113273}.

\bibitem[Team()]{teamgemini}
Gemini Team.
\newblock Gemini: A family of highly capable multimodal models.
\newblock Technical report, Technical report, Google, 12 2023. URL https://storage. googleapis. com~….

\bibitem[Touvron et~al.(2023{\natexlab{a}})Touvron, Lavril, Izacard, Martinet, Lachaux, Lacroix, Rozi{\`e}re, Goyal, Hambro, Azhar, et~al.]{touvron2023Llama}
Hugo Touvron, Thibaut Lavril, Gautier Izacard, Xavier Martinet, Marie-Anne Lachaux, Timoth{\'e}e Lacroix, Baptiste Rozi{\`e}re, Naman Goyal, Eric Hambro, Faisal Azhar, et~al.
\newblock Llama: Open and efficient foundation language models.
\newblock \emph{arXiv preprint arXiv:2302.13971}, 2023{\natexlab{a}}.

\bibitem[Touvron et~al.(2023{\natexlab{b}})Touvron, Martin, Stone, Albert, Almahairi, Babaei, Bashlykov, Batra, Bhargava, Bhosale, Bikel, Blecher, Ferrer, Chen, Cucurull, Esiobu, Fernandes, Fu, Fu, Fuller, Gao, Goswami, Goyal, Hartshorn, Hosseini, Hou, Inan, Kardas, Kerkez, Khabsa, Kloumann, Korenev, Koura, Lachaux, Lavril, Lee, Liskovich, Lu, Mao, Martinet, Mihaylov, Mishra, Molybog, Nie, Poulton, Reizenstein, Rungta, Saladi, Schelten, Silva, Smith, Subramanian, Tan, Tang, Taylor, Williams, Kuan, Xu, Yan, Zarov, Zhang, Fan, Kambadur, Narang, Rodriguez, Stojnic, Edunov, and Scialom]{Touvron2023Llama2O}
Hugo Touvron, Louis Martin, Kevin~R. Stone, Peter Albert, Amjad Almahairi, Yasmine Babaei, Nikolay Bashlykov, Soumya Batra, Prajjwal Bhargava, Shruti Bhosale, Daniel~M. Bikel, Lukas Blecher, Cristian~Cant{\'o}n Ferrer, Moya Chen, Guillem Cucurull, David Esiobu, Jude Fernandes, Jeremy Fu, Wenyin Fu, Brian Fuller, Cynthia Gao, Vedanuj Goswami, Naman Goyal, Anthony~S. Hartshorn, Saghar Hosseini, Rui Hou, Hakan Inan, Marcin Kardas, Viktor Kerkez, Madian Khabsa, Isabel~M. Kloumann, A.~V. Korenev, Punit~Singh Koura, Marie-Anne Lachaux, Thibaut Lavril, Jenya Lee, Diana Liskovich, Yinghai Lu, Yuning Mao, Xavier Martinet, Todor Mihaylov, Pushkar Mishra, Igor Molybog, Yixin Nie, Andrew Poulton, Jeremy Reizenstein, Rashi Rungta, Kalyan Saladi, Alan Schelten, Ruan Silva, Eric~Michael Smith, R.~Subramanian, Xia Tan, Binh Tang, Ross Taylor, Adina Williams, Jian~Xiang Kuan, Puxin Xu, Zhengxu Yan, Iliyan Zarov, Yuchen Zhang, Angela Fan, Melanie Kambadur, Sharan Narang, Aurelien Rodriguez, Robert Stojnic, Sergey Edunov, and
  Thomas Scialom.
\newblock Llama 2: Open foundation and fine-tuned chat models.
\newblock \emph{ArXiv}, abs/2307.09288, 2023{\natexlab{b}}.
\newblock URL \url{https://api.semanticscholar.org/CorpusID:259950998}.

\bibitem[Tunstall et~al.(2023)Tunstall, Beeching, Lambert, Rajani, Rasul, Belkada, Huang, von Werra, Fourrier, Habib, Sarrazin, Sanseviero, Rush, and Wolf]{tunstall2023zephyr}
Lewis Tunstall, Edward Beeching, Nathan Lambert, Nazneen Rajani, Kashif Rasul, Younes Belkada, Shengyi Huang, Leandro von Werra, Clémentine Fourrier, Nathan Habib, Nathan Sarrazin, Omar Sanseviero, Alexander~M. Rush, and Thomas Wolf.
\newblock Zephyr: Direct distillation of lm alignment, 2023.

\bibitem[Van~de Ven et~al.(2022)Van~de Ven, Tuytelaars, and Tolias]{van2022three}
Gido~M Van~de Ven, Tinne Tuytelaars, and Andreas~S Tolias.
\newblock Three types of incremental learning.
\newblock \emph{Nature Machine Intelligence}, 4\penalty0 (12):\penalty0 1185--1197, 2022.

\bibitem[Villalobos et~al.(2022)Villalobos, Sevilla, Heim, Besiroglu, Hobbhahn, and Ho]{Villalobos2022WillWR}
Pablo Villalobos, Jaime Sevilla, Lennart Heim, Tamay Besiroglu, Marius Hobbhahn, and An~Chang Ho.
\newblock Will we run out of data? an analysis of the limits of scaling datasets in machine learning.
\newblock \emph{ArXiv}, abs/2211.04325, 2022.
\newblock URL \url{https://api.semanticscholar.org/CorpusID:253397775}.

\bibitem[Welbl et~al.(2017)Welbl, Liu, and Gardner]{Welbl2017CrowdsourcingMC}
Johannes Welbl, Nelson~F. Liu, and Matt Gardner.
\newblock Crowdsourcing multiple choice science questions.
\newblock \emph{ArXiv}, abs/1707.06209, 2017.
\newblock URL \url{https://api.semanticscholar.org/CorpusID:1553193}.

\bibitem[Wu et~al.(2023)Wu, Zhang, Zhang, Wang, and Xie]{Wu2023PMCLlamaTB}
Chaoyi Wu, Xiaoman Zhang, Ya~Zhang, Yanfeng Wang, and Weidi Xie.
\newblock Pmc-llama: Towards building open-source language models for medicine.
\newblock 2023.
\newblock URL \url{https://api.semanticscholar.org/CorpusID:258417843}.

\bibitem[Wu et~al.(2022)Wu, Logan~IV, Walsh, Bhagia, Groeneveld, Singh, and Beltagy]{wu2022continued}
Zhaofeng Wu, Robert~L Logan~IV, Pete Walsh, Akshita Bhagia, Dirk Groeneveld, Sameer Singh, and Iz~Beltagy.
\newblock Continued pretraining for better zero-and few-shot promptability.
\newblock \emph{arXiv preprint arXiv:2210.10258}, 2022.

\bibitem[Xie et~al.(2024{\natexlab{a}})Xie, Chen, Chen, Peng, Hu, Lin, Peng, Huang, Zhang, Keloth, Zhou, He, Ohno-Machido, Wu, Xu, and Bian]{Xie2024MeLF}
Qianqian Xie, Qingyu Chen, Aokun Chen, C.A.I. Peng, Yan Hu, Fongci Lin, Xueqing Peng, Jimin Huang, Jeffrey Zhang, Vipina~Kuttichi Keloth, Xingyu Zhou, Huan He, Lucila Ohno-Machido, Yonghui Wu, Hua Xu, and Jiang Bian.
\newblock Me llama: Foundation large language models for medical applications.
\newblock \emph{ArXiv}, abs/2402.12749, 2024{\natexlab{a}}.
\newblock URL \url{https://api.semanticscholar.org/CorpusID:267759846}.

\bibitem[Xie et~al.(2024{\natexlab{b}})Xie, Chen, Chen, Peng, Hu, Lin, Peng, Huang, Zhang, Keloth, et~al.]{xie2024me}
Qianqian Xie, Qingyu Chen, Aokun Chen, Cheng Peng, Yan Hu, Fongci Lin, Xueqing Peng, Jimin Huang, Jeffrey Zhang, Vipina Keloth, et~al.
\newblock Me llama: Foundation large language models for medical applications.
\newblock \emph{arXiv preprint arXiv:2402.12749}, 2024{\natexlab{b}}.

\bibitem[Xie et~al.(2024{\natexlab{c}})Xie, Pham, Dong, Du, Liu, Lu, Liang, Le, Ma, and Yu]{xie2024doremi}
Sang~Michael Xie, Hieu Pham, Xuanyi Dong, Nan Du, Hanxiao Liu, Yifeng Lu, Percy~S Liang, Quoc~V Le, Tengyu Ma, and Adams~Wei Yu.
\newblock Doremi: Optimizing data mixtures speeds up language model pretraining.
\newblock \emph{Advances in Neural Information Processing Systems}, 36, 2024{\natexlab{c}}.

\bibitem[Xue et~al.(2023)Xue, Fu, Zhou, Zheng, and You]{Xue2023ToRO}
Fuzhao Xue, Yao Fu, Wangchunshu Zhou, Zangwei Zheng, and Yang You.
\newblock To repeat or not to repeat: Insights from scaling llm under token-crisis.
\newblock \emph{ArXiv}, abs/2305.13230, 2023.
\newblock URL \url{https://api.semanticscholar.org/CorpusID:258833284}.

\bibitem[Xue et~al.(2024)Xue, Fu, Zhou, Zheng, and You]{xue2024repeat}
Fuzhao Xue, Yao Fu, Wangchunshu Zhou, Zangwei Zheng, and Yang You.
\newblock To repeat or not to repeat: Insights from scaling llm under token-crisis.
\newblock \emph{Advances in Neural Information Processing Systems}, 36, 2024.

\bibitem[Yang et~al.(2024{\natexlab{a}})Yang, Gao, Xue, and Alexandersson]{yang2024pLlama}
Xianjun Yang, Junfeng Gao, Wenxin Xue, and Erik Alexandersson.
\newblock Pllama: An open-source large language model for plant science.
\newblock \emph{arXiv preprint arXiv:2401.01600}, 2024{\natexlab{a}}.

\bibitem[Yang et~al.(2024{\natexlab{b}})Yang, Cao, and Zhao]{Yang2024LaCoLL}
Yifei Yang, Zouying Cao, and Hai Zhao.
\newblock Laco: Large language model pruning via layer collapse.
\newblock \emph{ArXiv}, abs/2402.11187, 2024{\natexlab{b}}.
\newblock URL \url{https://api.semanticscholar.org/CorpusID:267751181}.

\bibitem[Y{\i}ld{\i}z et~al.(2024)Y{\i}ld{\i}z, Ravichandran, Punia, Bethge, and Ermis]{yildiz2024investigating}
{\c{C}}a{\u{g}}atay Y{\i}ld{\i}z, Nishaanth~Kanna Ravichandran, Prishruit Punia, Matthias Bethge, and Beyza Ermis.
\newblock Investigating continual pretraining in large language models: Insights and implications.
\newblock \emph{arXiv preprint arXiv:2402.17400}, 2024.

\bibitem[Zellers et~al.(2019)Zellers, Holtzman, Bisk, Farhadi, and Choi]{zellers2019hellaswag}
Rowan Zellers, Ari Holtzman, Yonatan Bisk, Ali Farhadi, and Yejin Choi.
\newblock Hellaswag: Can a machine really finish your sentence?
\newblock \emph{arXiv preprint arXiv:1905.07830}, 2019.

\bibitem[Zhang et~al.(2023{\natexlab{a}})Zhang, Yu, Yan, Liu, Adhikarla, Fu, Chen, Chen, Zhou, Li, et~al.]{zhang2023biomedgpt}
Kai Zhang, Jun Yu, Zhiling Yan, Yixin Liu, Eashan Adhikarla, Sunyang Fu, Xun Chen, Chen Chen, Yuyin Zhou, Xiang Li, et~al.
\newblock Biomedgpt: A unified and generalist biomedical generative pre-trained transformer for vision, language, and multimodal tasks.
\newblock \emph{arXiv preprint arXiv:2305.17100}, 2023{\natexlab{a}}.

\bibitem[Zhang et~al.(2024{\natexlab{a}})Zhang, Zeng, Wang, and Lu]{Zhang2024TinyLlamaAO}
Peiyuan Zhang, Guangtao Zeng, Tianduo Wang, and Wei Lu.
\newblock Tinyllama: An open-source small language model.
\newblock \emph{ArXiv}, abs/2401.02385, 2024{\natexlab{a}}.
\newblock URL \url{https://api.semanticscholar.org/CorpusID:266755802}.

\bibitem[Zhang et~al.(2024{\natexlab{b}})Zhang, Zeng, Wang, and Lu]{zhang2024tinyLlama}
Peiyuan Zhang, Guangtao Zeng, Tianduo Wang, and Wei Lu.
\newblock Tinyllama: An open-source small language model, 2024{\natexlab{b}}.

\bibitem[Zhang et~al.(2023{\natexlab{b}})Zhang, Tian, Yang, Chen, Li, and Petzold]{zhang2023alpacareinstructiontuned}
Xinlu Zhang, Chenxin Tian, Xianjun Yang, Lichang Chen, Zekun Li, and Linda~Ruth Petzold.
\newblock Alpacare:instruction-tuned large language models for medical application, 2023{\natexlab{b}}.

\bibitem[Zhang et~al.(2024{\natexlab{c}})Zhang, Luo, Yuan, and Yao]{Zhang2024AutoMathTextAD}
Yifan Zhang, Yifan Luo, Yang Yuan, and Andrew Chi-Chih Yao.
\newblock Automathtext: Autonomous data selection with language models for mathematical texts.
\newblock \emph{ArXiv}, abs/2402.07625, 2024{\natexlab{c}}.
\newblock URL \url{https://api.semanticscholar.org/CorpusID:267627801}.

\bibitem[Zhong et~al.(2023)Zhong, Cui, Guo, Liang, Lu, Wang, Saied, Chen, and Duan]{zhong2023agieval}
Wanjun Zhong, Ruixiang Cui, Yiduo Guo, Yaobo Liang, Shuai Lu, Yanlin Wang, Amin Saied, Weizhu Chen, and Nan Duan.
\newblock Agieval: A human-centric benchmark for evaluating foundation models.
\newblock \emph{arXiv preprint arXiv:2304.06364}, 2023.

\bibitem[Zhou et~al.(2023)Zhou, Xu, Zhu, Zhou, Lo, Sridhar, Cheng, Bisk, Fried, Alon, et~al.]{zhou2023webarena}
Shuyan Zhou, Frank~F Xu, Hao Zhu, Xuhui Zhou, Robert Lo, Abishek Sridhar, Xianyi Cheng, Yonatan Bisk, Daniel Fried, Uri Alon, et~al.
\newblock Webarena: A realistic web environment for building autonomous agents.
\newblock \emph{arXiv preprint arXiv:2307.13854}, 2023.
\newblock URL \url{https://webarena.dev}.

\bibitem[Zhuang et~al.(2024)Zhuang, Zhang, Zheng, Du, Wang, Ren, Huang, Fu, Yue, and Chen]{zhuang2024structlm}
Alex Zhuang, Ge~Zhang, Tianyu Zheng, Xinrun Du, Junjie Wang, Weiming Ren, Stephen~W Huang, Jie Fu, Xiang Yue, and Wenhu Chen.
\newblock Structlm: Towards building generalist models for structured knowledge grounding.
\newblock \emph{arXiv preprint arXiv:2402.16671}, 2024.

\end{thebibliography}


\appendix
\section{The Details of Pre-training}
\label{appendix.train_details_1}
For OpenLLaMa-3B, TinyLLaMa-1B, and Pythia-410m, we download them from their official website. For OpenLLaMa-3B and TinyLLaMa-1B LLMs, we continually pre-train them with the 50 billion medical tokens we constructed in Sec.~\ref{sec.method} for one epoch. For the Pythia-410m LLM, we continually pre-train it with the 100 billion tokens randomly sampled from the 2021-2022 subset of the RefinedWeb dataset. We consider this subset as the Pile dataset only contains data before the year 2021 and then the tokens sampled from the 2021-2022 subset are unseen for the Pythia-410m model. The pre-training code is based on the transformers. The task is to predict the next token with a context size of 2048. The training is executed using 192 V100 GPUs. We employ the AdamW optimizer with $\beta_1=0.9,\beta_2=0.95$,
a weight decay of 0.01, and a learning rate of 3e-4. We use a cosine learning rate scheduler with a 0.1 warmup ratio for gradual adaptation to training complexity and bf16 precision for computational efficiency. Gradient accumulation is set to 4 steps, and each training batch contains about 340 million tokens. We also add support for FlashAttention-2 ~\citep{dao2023flashattention} for more efficient inference and long-context decoding. 
\section{Task Information}
\label{appendix.task}
For the medical evaluation, we follow~\cite{chen2023meditron} and mainly consider the following four tasks:

\textbf{MedMCQA}~\citep{pmlr-v174-pal22a} is
a large-scale and comprehensive dataset for multichoice (four-option) medical question answering. It is derived from real-world medical entrance exam questions (Indian AIIMS and NEET-PG) and consists of over 194,000 high-quality medical questions. These questions cover 2,400 healthcare topics and 21 medical subjects, exhibiting a wide range
of topical diversity. The average token length is
12.77.

\textbf{MedQA}~\citep{jin2021disease}is a multichoice question-answering dataset collected from
the professional medical board exam, the United States Medical License Exams (USMLE). It comprises 12,723 questions sourced from a comprehensive collection of 18 English medical textbooks that
have been extensively utilized by medical students
and USMLE candidates. Questions in MedQA
cover a wide range of topics in clinical medicine,
necessitating responses with professional expertise
and complex multi-hop reasoning across multiple
pieces of evidence. The average question and option length is 116.6 and 3.5, respectively.

\textbf{MMLU}~\citep{Hendrycks2020MeasuringMM} is a comprehensive multi-task language understanding
test dataset that encompasses 57 tasks across various domains such as mathematics, history, computer science, law, etc. In our experiments,
we specifically focus on a subset of medical reasoning-related tasks including clinical knowledge, college medicine, medical genetics, and professional medicine.

\textbf{PubMedQA}~\citep{jin2019pubmedqa} is
a biomedical question and answering dataset derived from PubMed abstracts. It contains 1k expert annotated multi-choice question-and-answer samples based on 211.3k PubMed articles. The task
of PubMedQA is to provide answers to research
questions with yes/no/maybe responses based on
the corresponding abstracts. The average question
and context length is 14.4 and 238.9, respectively.

\textbf{HOC}~\citep{baker2016automatic} is a classification task to decide the Hallmarks of Cancer (HOC) taxonomy of the article based on its abstract. The input is an abstract text. There are 10 topics you will need to decide whether the article is related to. Topics:
sustaining proliferative signaling, evading growth suppressors, resisting cell death, enabling replicative
immortality, inducing angiogenesis, activating invasion and metastasis, genomic instability and mutation, tumor
promoting inflammation, and cellular energetics, and avoiding immune destruction. 

\textbf{DDI 2023}~\citep{segura2013semeval} is a task to predict the relationship between the given head entity labeled as @DRUG1$ and tail entity labeled
as @DRUG2$ within a given sentence, this relation which must be in (‘mechanism’, ‘effect’, ‘advice’, ‘int’,
'none’). mechanism: this type is used to annotate drug-drug interactions that are described by their
pharmacokinetic mechanism. effect: this type is used to annotate drug-drug interactions describing an effect or
a pharmacodynamic mechanism. advice: this type is used when a recommendation or advice regarding a drug
interaction is given. int: this type is used when a drug-drug interaction appears in the text without providing
any additional information. none: there are no drug-drug interactions. 

\textbf{BioNLI}~\citep{bastan2022bionli} is a task to classify the relationship between the given medical premise and hypothesis into one of the following labels:
entailment, contradiction, or neutral. This dataset contains abstracts from biomedical literature and mechanistic premises generated with nine different strategies.

\textbf{MIMIC-CXR}~\citep{johnson2019mimic} is a generation task that derives the impression from findings in the radiology report.

The dataset statistics are in Table~\ref{tab.statistics}
\begin{table}[h!]
\centering
\caption{Dataset statistics}
\begin{tabular}{lccc}
\toprule
\textbf{Dataset} & \textbf{\# Train} & \textbf{\# Test} & \textbf{Source} \\
\midrule
{MedMCQA~\citep{pmlr-v174-pal22a}} & 182,822 & 4183 & Exam \\
{MedQA~\citep{jin2021disease}} & 10178 & 1273 & Exam \\
{MMLU~\citep{Hendrycks2020MeasuringMM}} & - & 163 & Exam \\
{PubMedQA~\citep{jin2019pubmedqa}} & 211,269 & 500 & Literature \\
{HOC~\citep{baker2016automatic}} & 1108 & 315 & Literature \\
{DDI 2023~\citep{segura2013semeval}} & 1108 & 315 & Literature \\
{BioNLI~\citep{bastan2022bionli}} & 5544 & 6308 & Literature \\
{MIMIC-CXR~\citep{bastan2022bionli}} & 122,014 & 1606 & Literature \\
\bottomrule
\end{tabular}
\label{tab.statistics}
\end{table}

For the evaluation of general task ability, we consider the following 10 commonsense tasks:

\textbf{ARC-Challenge and ARC-Easy} ARC~\citep{clark2018think} is a multiple-choice question-answering dataset, containing questions from science exams from grade 3 to grade 9. The dataset is split into two partitions: Easy and Challenge, where the latter partition contains the more difficult questions that require reasoning. Most of the questions have 4 answer choices.

\textbf{BoolQ}~\citep{clark2019boolq} is a question-answering dataset for yes/no questions containing 15942 examples. These questions are naturally occurring ---they are generated in unprompted and unconstrained settings. Each example is a triplet of (question, passage, answer), with the title of the page as optional additional context. The text-pair classification setup is similar to existing natural language inference tasks.

\textbf{COPA}~\citep{roemmele2011choice} consists of 1000 questions, split equally into development and test sets of 500 questions each. Each question is composed of a premise and two alternatives, where the task is to select the alternative that more plausibly has a causal relation with the premise. 

\textbf{HellaSWAG}~\citep{zellers2019hellaswag} is a dataset for studying grounded commonsense inference. It consists of 70k multiple choice questions about grounded situations: each question comes from one of two domains -- activitynet or wikihow -- with four answer choices about what might happen next in the scene. The correct answer is the (real) sentence for the next event; the three incorrect answers are adversarially generated and human-verified, so as to fool machines but not humans.

\textbf{OpenBookQA}~\citep{Mihaylov2018CanAS} is a new kind of question-answering dataset modeled after open-book exams for assessing human understanding of a subject. It consists of 5,957 multiple-choice elementary-level science questions (4,957 train, 500 dev, 500 test), which probe the understanding of a small “book” of 1,326 core science facts and the application of these facts to novel situations.

\textbf{PIQA}~\citep{bisk2020piqa} dataset introduces the task of physical commonsense reasoning and a corresponding benchmark dataset Physical Interaction: Question Answering or PIQA. Physical commonsense knowledge is a major challenge on the road to true AI-completeness, including robots that interact with the world and understand natural language. PIQA focuses on everyday situations with a preference for atypical solutions. 

\textbf{Race}~\citep{lai2017large} is a large-scale reading comprehension dataset with more than 28,000 passages and nearly 100,000 questions. The dataset is collected from English examinations in China, which are designed for middle school and high school students. The dataset can serve as the training and test sets for machine comprehension.

\textbf{SciQ}~\citep{Welbl2017CrowdsourcingMC} dataset contains 13,679 crowdsourced science exam questions about Physics, Chemistry and Biology, among others. The questions are in multiple-choice format with 4 answer options each. For the majority of the questions, an additional paragraph with supporting evidence for the correct answer is provided.

\textbf{WinoGrande}~\citep{ai2:winogrande} is a new collection of 44k problems, inspired by the Winograd Schema Challenge (Levesque, Davis, and Morgenstern 2011), but adjusted to improve the scale and robustness against the dataset-specific bias. Formulated as a fill-in-a-blank task with binary options, the goal is to choose the right option for a given sentence which requires commonsense reasoning.

We use the lm-eval-harness~\citep{eval-harness} to evaluate the LLM on these tasks' test set and report the zero-shot performance.

\section{The Perplexity and relative parameter update rate of the LLM using our strategies}
\label{appendix:ppl}
\begin{figure*}[htp]
\centering 
\includegraphics[height=0.35\textwidth,width=0.95\textwidth]{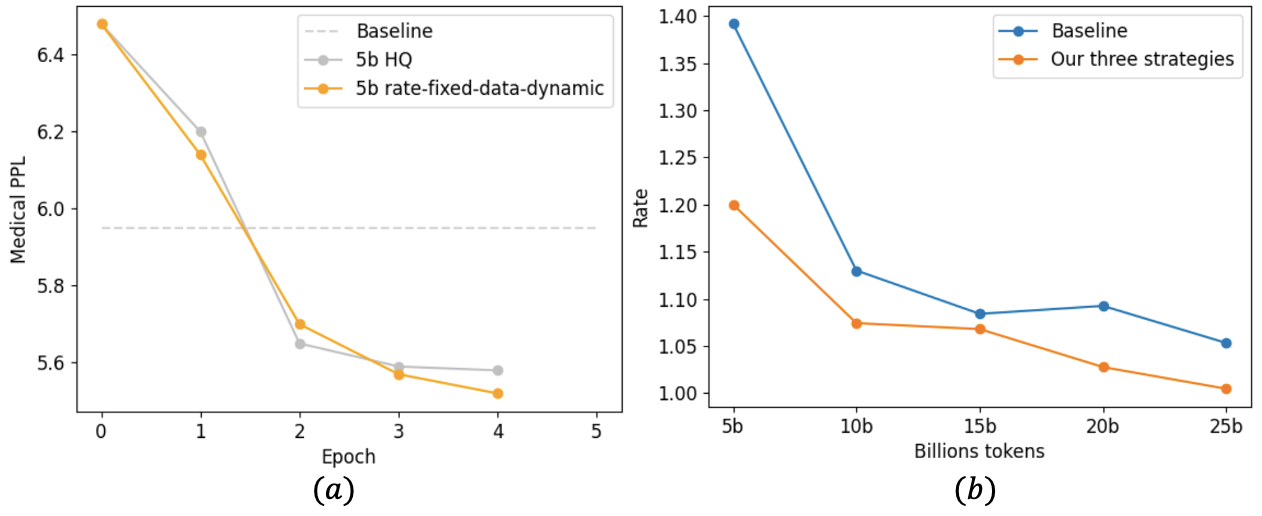} 
\vspace{-3mm}
\caption{(a) reports the average medical perplexity of the OpenLLaMa-3B using our strategies. '5b HQ' means the LLM using our strategies I and II. '5b rate-fixed-data-dynamic' means the LLM using our three strategies. 'Baseline' is the average medical perplexity of the OpenLLaMa-3B model that has been continually pre-trained with 50 billion medical tokens. (b) shows the rate between the bottom 5 layers' average relative parameter and the top 5 layers' average relative parameter update of the OpenLLaMa-3B using our strategies. 'Baseline' is the rate of the OpenLLaMa-3B model during the continual pre-training with 50 billion medical tokens. 
}
\label{Fig.analysis_4}
\vspace{-3mm}
\end{figure*}
From Figure~\ref{Fig.analysis_4}(a), we observe that the LLM using our strategies gradually decreases its average medical perplexity, indicating that the LLM is acquiring rich medical knowledge. Its average medical perplexity at the fourth epoch is even lower than that of the OpenLLaMa-3B model, which has been continually pre-trained with 50 billion medical tokens. From Figure~\ref{Fig.analysis_4}(b), we also find that the ratio between the average relative parameter updates of the bottom 5 layers and the top 5 layers of the OpenLLaMa-3B model using our strategies is closer to 1. This suggests that the plasticity gradient and the stability gradient are more balanced when employing our strategies.

\section{Deploying our strategies into the general continual pre-training setting}
\label{appendix.general}
\begin{figure*}[htp]
\centering 
\includegraphics[height=0.35\textwidth,width=0.5\textwidth]{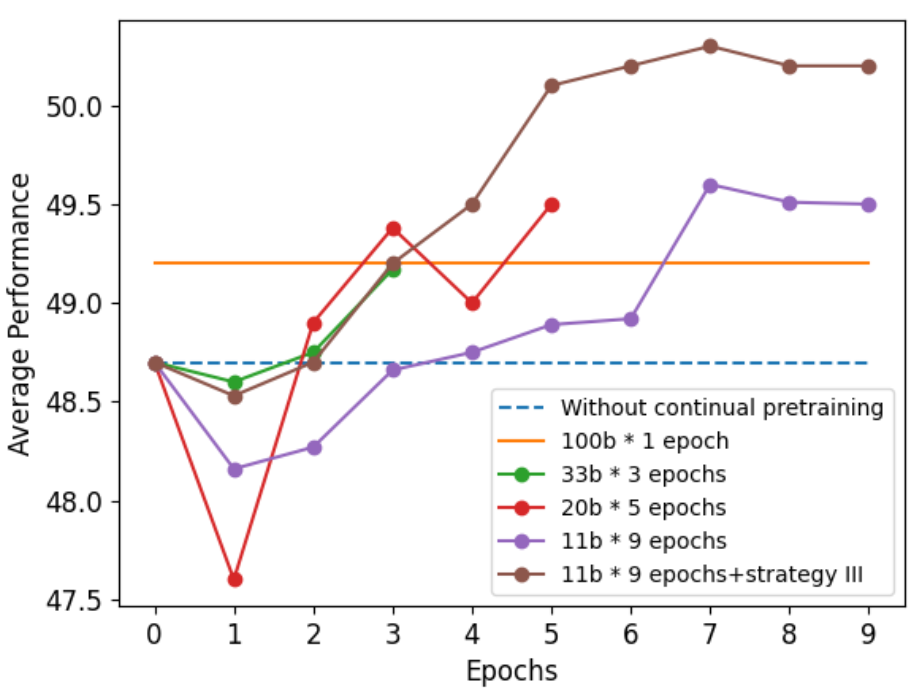} 
\vspace{-3mm}
\caption{We report the average performance of the 10 commonsense and reading compression task here. The Model is Pythia-410m. 
}
\label{Fig.analysis_5}
\vspace{-3mm}
\end{figure*}

Continually pre-training one LLM on another large corpus is an approach to boost its general ability \citep{Gupta2023ContinualPO}. We consider the scenario of continually pre-training the Pythia-410m model on the RefinedWeb dataset. The Pythia-410m model has been pre-trained on the Pile dataset. In this context, we use the average performance of 10 commonsense and reading comprehension tasks, as detailed in Appendix~\ref{appendix.task}, to measure the LLM's general task performance. To test the effectiveness of strategy I in the general continual pre-training setting, we conduct multi-epoch experiments with different training subset sizes. The tokens in each training subset are randomly sampled from the RefinedWeb dataset and the computational consumption of each experiment can not be beyond the compute budget (100 billion tokens). From Figure~\ref{Fig.analysis_5}, we find that strategy I indeed helps the Pythia-410m model to mitigate the stability gap and achieve better peak performance. We also find the best performance among our experiments is achieved when pre-training the LLM with 11 billion tokens for 7 epochs. However, we can not find a good quality filter for the second strategy. We have tried to train a KenLM on WikiText as the quality filter for measuring the sample's quality in improving LLMs' general ability. But it does not work. From Figure~\ref{Fig.analysis_5}, we find that strategies I and III can help the LLM to overcome the stability gap and achieve higher performance.
\section{The Training Details of Deploying our Strategies into the Llama-3 Model}
\label{appendix.llama-3}

\textbf{Pre-training details:} The pre-training task is to predict the next token with a context size of 8192. The training is executed using 16 H100 80GB GPUs. We employ the AdamW optimizer with $\beta_1=0.9,\beta_2=0.95$,
a weight decay of 0.01, and a learning rate of 3e-5. We use a cosine learning rate scheduler with a 0.1 warmup ratio for gradual adaptation to training complexity and bf16 precision for computational efficiency. Gradient accumulation is set to 12 steps, and each training batch contains about 340 million tokens. We also add support for FlashAttention-2 ~\citep{dao2023flashattention} for more efficient inference and long-context decoding.

\textbf{Task-specific finetuning details:} 
We employ the AdamW optimizer with a weight decay of 0.01 and a learning rate of 3e-5. We use a cosine learning rate schedule with a 10\% warmup ratio, decaying the final learning rate to 10\% of the peak learning rate. We fine-tune the LLMs for 3 epochs.

\textbf{Instructions-tuning details:} We consider the combination of the training set of MedMCQA~\citep{pmlr-v174-pal22a}, MedQA~\citep{jin2021disease}, PubMedQA~\citep{jin2019pubmedqa}, HOC~\citep{baker2016automatic}, DDI2013~\citep{segura2013semeval}, BioNLI~\citep{bastan2022bionli}, and MIMIC-CXR~\citep{johnson2019mimic} tasks . To avoid potential data contamination, for each test sample of  MedQA~\citep{jin2021disease}, PubMedQA~\citep{jin2019pubmedqa}, and MedMCQA~\citep{pmlr-v174-pal22a} tasks, we delete the training samples that contain its option. For the training samples of theMedQA~\citep{jin2021disease},PubMedQA~\citep{jin2019pubmedqa}, and MedMCQA~\citep{pmlr-v174-pal22a} tasks, we use the instruction template from the Meditron paper~\citep{chen2023meditron}. For the other datasets' training samples, we use their original instructions. We employ the AdamW optimizer with a weight decay of 0.01 and a learning rate of 3e-5. We use a cosine learning rate schedule with a 10\% warmup ratio, decaying the final learning rate to 10\% of the peak learning rate. We fine-tune the LLMs for 3 epochs. The global batch size is 96 and max sequence length is 1024.

\newpage
\section*{NeurIPS Paper Checklist}

The checklist is designed to encourage best practices for responsible machine learning research, addressing issues of reproducibility, transparency, research ethics, and societal impact. Do not remove the checklist: {\bf The papers not including the checklist will be desk rejected.} The checklist should follow the references and precede the (optional) supplemental material.  The checklist does NOT count towards the page
limit. 

Please read the checklist guidelines carefully for information on how to answer these questions. For each question in the checklist:
\begin{itemize}
    \item You should answer \answerYes{}, \answerNo{}, or \answerNA{}.
    \item \answerNA{} means either that the question is Not Applicable for that particular paper or the relevant information is Not Available.
    \item Please provide a short (1–2 sentence) justification right after your answer (even for NA). 
\end{itemize}

{\bf The checklist answers are an integral part of your paper submission.} They are visible to the reviewers, area chairs, senior area chairs, and ethics reviewers. You will be asked to also include it (after eventual revisions) with the final version of your paper, and its final version will be published with the paper.

The reviewers of your paper will be asked to use the checklist as one of the factors in their evaluation. While "\answerYes{}" is generally preferable to "\answerNo{}", it is perfectly acceptable to answer "\answerNo{}" provided a proper justification is given (e.g., "error bars are not reported because it would be too computationally expensive" or "we were unable to find the license for the dataset we used"). In general, answering "\answerNo{}" or "\answerNA{}" is not grounds for rejection. While the questions are phrased in a binary way, we acknowledge that the true answer is often more nuanced, so please just use your best judgment and write a justification to elaborate. All supporting evidence can appear either in the main paper or the supplemental material, provided in appendix. If you answer \answerYes{} to a question, in the justification please point to the section(s) where related material for the question can be found.

IMPORTANT, please:
\begin{itemize}
    \item {\bf Delete this instruction block, but keep the section heading ``NeurIPS paper checklist"},
    \item  {\bf Keep the checklist subsection headings, questions/answers and guidelines below.}
    \item {\bf Do not modify the questions and only use the provided macros for your answers}.
\end{itemize}


\begin{enumerate}

\item {\bf Claims}
    \item[] Question: Do the main claims made in the abstract and introduction accurately reflect the paper's contributions and scope?
    \item[] Answer: \answerYes{} 
    \item[] Justification: Our abstract and introduction clearly state the claims made, including the contributions made in the paper.

\item {\bf Limitations}
    \item[] Question: Does the paper discuss the limitations of the work performed by the authors?
    \item[] Answer: \answerYes{} 
    \item[] Justification: Yes, we discuss our limitations after the conclusion section. 

\item {\bf Theory Assumptions and Proofs}
    \item[] Question: For each theoretical result, does the paper provide the full set of assumptions and a complete (and correct) proof?
    \item[] Answer: \answerNA{} 
    \item[] Justification: We focus on the behavior of LLMs during continual pre-training. \justificationTODO{}

    \item {\bf Experimental Result Reproducibility}
    \item[] Question: Does the paper fully disclose all the information needed to reproduce the main experimental results of the paper to the extent that it affects the main claims and/or conclusions of the paper (regardless of whether the code and data are provided or not)?
    \item[] Answer: \answerYes{} 
    \item[] Justification: We list all experimental details in the evaluation section and appendixes. We will release our models in public (e.g., HuggingFace). 

\item {\bf Open access to data and code}
    \item[] Question: Does the paper provide open access to the data and code, with sufficient instructions to faithfully reproduce the main experimental results, as described in supplemental material?
    \item[] Answer: \answerYes{} 
    \item[] Justification: The pre-training data is too large to put it into the material. The fine-tuning data is open-source. We will release the code and relevant data to reproduce our results soon. 

\item {\bf Experimental Setting/Details}
    \item[] Question: Does the paper specify all the training and test details (e.g., data splits, hyperweights, how they were chosen, type of optimizer, etc.) necessary to understand the results?
    \item[] Answer: \answerYes{} 
    \item[] Justification: We write all the training and test details in the evaluation section and appendixes.

\item {\bf Experiment Statistical Significance}
    \item[] Question: Does the paper report error bars suitably and correctly defined or other appropriate information about the statistical significance of the experiments?
    \item[] Answer: \answerNo{} 
    \item[] Justification: \justificationTODO{}

\item {\bf Experiments Compute Resources}
    \item[] Question: For each experiment, does the paper provide sufficient information on the computer resources (type of compute workers, memory, time of execution) needed to reproduce the experiments?
    \item[] Answer: \answerYes{} 
    \item[] Justification: See the evaluation section and appendixes.
    
\item {\bf Code Of Ethics}
    \item[] Question: Does the research conducted in the paper conform, in every respect, with the NeurIPS Code of Ethics \url{https://neurips.cc/public/EthicsGuidelines}?
    \item[] Answer: \answerYes{} 
    \item[] Justification: We conform with the NeurIPS Code of Ethics. 

\item {\bf Broader Impacts}
    \item[] Question: Does the paper discuss both potential positive societal impacts and negative societal impacts of the work performed?
    \item[] Answer: \answerYes{} 
    \item[] Justification: Yes, please read the paragraph after the conclusion section. 
    
\item {\bf Safeguards}
    \item[] Question: Does the paper describe safeguards that have been put in place for responsible release of data or models that have a high risk for misuse (e.g., pre-trained language models, image generators, or scraped datasets)?
    \item[] Answer: \answerYes{} 
    \item[] Justification: Yes, we describe it in the paragraph after the conclusion section. 

\item {\bf Licenses for existing assets}
    \item[] Question: Are the creators or original owners of assets (e.g., code, data, models), used in the paper, properly credited and are the license and terms of use explicitly mentioned and properly respected?
    \item[] Answer: \answerYes{} 
    \item[] Justification: Yes, we properly credit them. 

\item {\bf New Assets}
    \item[] Question: Are new assets introduced in the paper well documented and is the documentation provided alongside the assets?
    \item[] Answer: \answerYes{} 
    \item[] Justification: Yes, we introduce the details of our new models in the main paper. 

\item {\bf Crowdsourcing and Research with Human Subjects}
    \item[] Question: For crowdsourcing experiments and research with human subjects, does the paper include the full text of instructions given to participants and screenshots, if applicable, as well as details about compensation (if any)? 
    \item[] Answer: \answerNA{} 
    \item[] Justification: The paper does not involve crowdsourcing nor research with human subjects.

\item {\bf Institutional Review Board (IRB) Approvals or Equivalent for Research with Human Subjects}
    \item[] Question: Does the paper describe potential risks incurred by study participants, whether such risks were disclosed to the subjects, and whether Institutional Review Board (IRB) approvals (or an equivalent approval/review based on the requirements of your country or institution) were obtained?
    \item[] Answer: \answerNA{} 
    \item[] Justification: The paper does not involve crowdsourcing nor research with human subjects.

\end{enumerate}

\end{document}